\documentclass[sigconf]{aamas} 

\usepackage{balance} 
\usepackage{svg}
\usepackage{threeparttable}
\usepackage{wrapfig}

\newcommand{\etal}{\textit{et al}. }

\setcopyright{ifaamas}
\acmConference[AAMAS '23]{Proc.\@ of the 22nd International Conference
on Autonomous Agents and Multiagent Systems (AAMAS 2023)}{May 29 -- June 2, 2023}
{London, United Kingdom}{A.~Ricci, W.~Yeoh, N.~Agmon, B.~An (eds.)}
\copyrightyear{2023}
\acmYear{2023}
\acmDOI{}
\acmPrice{}
\acmISBN{}

\acmSubmissionID{1180}

\title[]{Structural Attention-Based Recurrent Variational Autoencoder\\ for Highway Vehicle Anomaly Detection}

\author{Neeloy Chakraborty}
\affiliation{
  \institution{University of Illinois}
  \city{Urbana, IL}
  \country{United States}
  }
\email{neeloyc2@illinois.edu}

\author{Aamir Hasan*}
\affiliation{
  \institution{University of Illinois}
  \city{Urbana, IL}
  \country{United States}
  }
\email{aamirh2@illinois.edu}

\author{Shuijing Liu*}
\affiliation{
  \institution{University of Illinois}
  \city{Urbana, IL}
  \country{United States}
  }
\email{sliu105@illinois.edu}

\author{Tianchen Ji*}
\affiliation{
  \institution{University of Illinois}
  \city{Urbana, IL}
  \country{United States}
  }
\email{tj12@illinois.edu}

\author{Weihang Liang}
\affiliation{
  \institution{University of Illinois}
  \city{Urbana, IL}
  \country{United States}
  }
\email{weihang2@illinois.edu}

\author{D. Livingston McPherson}
\affiliation{
  \institution{University of Illinois}
  \city{Urbana, IL}
  \country{United States}
  }
\email{dlivm@illinois.edu}

\author{Katherine Driggs-Campbell}
\affiliation{
  \institution{University of Illinois}
  \city{Urbana, IL}
  \country{United States}
  }
\email{krdc@illinois.edu}

\begin{abstract}
In autonomous driving, detection of abnormal driving behaviors is essential to ensure the safety of vehicle controllers. 
Prior works in vehicle anomaly detection have shown that modeling interactions between agents improves detection accuracy, but certain abnormal behaviors where structured road information is paramount are poorly identified, such as wrong-way and off-road driving.
We propose a novel unsupervised framework for highway anomaly detection named Structural Attention-Based Recurrent VAE (SABeR-VAE), which explicitly uses the structure of the environment to aid anomaly identification. 
Specifically, we use a vehicle self-attention module to learn the relations among vehicles on a road, and a separate lane-vehicle attention module to model the importance of permissible lanes to aid in trajectory prediction. 
Conditioned on the attention modules' outputs, a recurrent encoder-decoder architecture with a stochastic Koopman operator-propagated latent space predicts the next states of vehicles.
Our model is trained end-to-end to minimize prediction loss on normal vehicle behaviors, and is deployed to detect anomalies in (ab)normal scenarios. 
By combining the heterogeneous vehicle and lane information, SABeR-VAE and its deterministic variant, SABeR-AE, improve abnormal AUPR by $18\%$ and $25\%$ respectively on the simulated MAAD highway dataset over STGAE-KDE.
Furthermore, we show that the learned Koopman operator in SABeR-VAE enforces interpretable structure in the variational latent space.
The results of our method indeed show that modeling environmental factors is essential to detecting a diverse set of anomalies in deployment. 
For code implementation, please visit \href{https://sites.google.com/illinois.edu/saber-vae}{\color{blue}{https://sites.google.com/illinois.edu/saber-vae}}.
\end{abstract}


\newcommand{\BibTeX}{\rm B\kern-.05em{\sc i\kern-.025em b}\kern-.08em\TeX}

\setcopyright{none}
\renewcommand\footnotetextcopyrightpermission[1]{} 

\begin{document}


\pagestyle{fancy}
\fancyhead{}


\maketitle 

\def\thefootnote{*}\footnotetext{These authors contributed equally to this work.}

\section{Introduction}
\label{sec:intro}
Autonomous vehicles have the potential to realize a fast, safe, and labor-free transportation system. 
A trustworthy self-driving vehicle should have the ability to operate reliably in normal situations and, more importantly, to perceive and react to anomalous driving scenarios (e.g., skidding and wrong-way driving of surrounding human vehicles) promptly and robustly. 
The detection of such abnormal situations can help identify traffic accidents and dangerous driving behaviors of road participants, and thus provide high-level guidance for vehicle controllers to act safely.

Deep-learning based Anomaly Detection (AD) algorithms have shown great promise in intelligent vehicle applications~\cite{bogdoll2022anomaly}.
Many previous works utilize vehicle trajectories as an anomaly signal~\cite{YangDrivingBeh, azzalini2020hmms, chen2021dual}. 
However, only a few vehicle trajectory datasets with sufficient anomaly labels exist for supervised learning methods~\cite{Zhang2021FindingCS, supvssemisup, YangDrivingBeh}. 
To leverage the larger store of unlabeled driving data, researchers like Yao and Wiederer have employed unsupervised learning methods~\cite{yao2019unsupervised,yao2022dota,MAAD2021}.
Specifically, a neural network, which generally follows an encoder-decoder architecture for trajectory reconstruction or prediction, learns an underlying distribution of normal vehicle trajectories in the latent space. 
An anomaly is then detected whenever the trajectory is out of distribution and produces a large reconstruction or prediction error. 
In interactive driving scenarios, Wiederer \etal ~\cite{MAAD2021} showed that modeling interactions between agents can largely improve the reconstruction accuracy and subsequently the AD performance. 
However, such interaction-aware methods still ignore the effect of road structures on vehicle behaviors, and thus can miss abnormal scenarios like wrong-way driving trajectories that appear normal when environmental context is overlooked.


Alongside performance accuracy, the decisions made by AD algorithms need to be interpretable to stakeholders. 
Deep neural networks are black boxes by nature. However, the decisions of deep networks impact various stakeholders such as policy makers and end users.
Designing methods with \emph{interpretable} features for stakeholders is a key challenge in AD, and the field of machine learning overall~\cite{Candido2021, sipple2022general, bhatt2020machine, SEJR2021}. 
In vehicle AD more specifically, interpretable algorithms need to account for the wide distribution of human drivers who act according to their own policies~\cite{bhattacharyya2019simulating}. 
For example, different drivers may choose to overtake other vehicles at different times and speeds.
To ensure interpretability, we use variational autoencoder (VAE) to cluster useful features from similar behaviors together in a continuous and stochastic latent space~\cite{kingma2014vae}.
Our results indicate that vehicle trajectories transitioning to an abnormal state are explicitly encoded by interpretable transformations in the learned latent space.

In this paper, we present our novel unsupervised Structural Attention-Based Recurrent Variational Autoencoder (SABeR-VAE) for highway vehicle anomaly detection.
Since contemporary vehicles have map information available to them regarding their nearby environment and lanes, we make use of the environmental information that prior works~\cite{MAAD2021, liu2022learning, scholler2020constant} have ignored to explicitly model the effect of lane structure on normal vehicle behaviors.
Specifically, we treat a highway scenario as a structured interaction graph where nodes represent vehicles and lane positions, and edges connect nearby vehicles, and permissible lanes.
Two separate attention modules learn relations between vehicles (vehicle-vehicle self-attention) and legal permissible route trajectories (lane-vehicle attention) respectively. 
A sequence of embeddings from the vehicle-vehicle attention module are encoded into a Gaussian latent space to capture the randomness of vehicle trajectories with a recurrent network, and cluster similar behaviors close together in an interpretable fashion.
Our work is more computationally efficient than STGAE-KDE~\cite{MAAD2021}, which has a deterministic latent space and requires the expensive process of fitting a Kernel Density Estimator (KDE) to learn a meaningful distribution of normal behaviors.
We then use a learned Koopman operator to propagate the current latent distributions forward in time conditioned on the useful lane embeddings.
We show that the Koopman operator explicitly enforces interpretable transformations in the latent space that standard autoencoders like STGAE are unable to incorporate, and is able to model the complex, non-linear dynamics of drivers.
Finally, we decode a sampled point from the propagated distribution to predict next states of vehicles.
We train our method to predict trajectories from normal scenarios in the Multi-Agent Anomaly Detection (MAAD) dataset~\cite{MAAD2021}, and compare accuracy metrics against linear, recurrent, and graph convolutional approaches on anomalous trajectories~\cite{scholler2020constant, park2018multimodal, MAAD2021}. Our SABeR-VAE improves AUPR-Abnormal and wrong-way driving detection over the STGAE-KDE by $18\%$ and $35\%$ respectively, and has an interpretable latent space.

Our contributions can be summarized as follows: 
(1) We present a novel unsupervised variational approach for anomaly detection conditioned on structured lane information; 
(2) We quantitatively show that incorporating the structured information increases anomaly detection accuracy, compared with state-of-the-art baselines and ablations using the MAAD dataset; 
(3) We show that the stochastic Koopman operator learns interpretable features of (ab)normal behaviors in the latent space.

Our paper is organized as follows: 
Section~\ref{sec:related} discusses relevant works in the areas of structured modeling and anomaly detection.
Our problem formulation and methods are presented in Section~\ref{sec:methods}.
We discuss results in Section~\ref{sec:results}.
Finally, we conclude the paper and discuss future directions in Section~\ref{sec:conclusion}.


\begin{figure}
    \centering
    \includegraphics[width=\columnwidth]{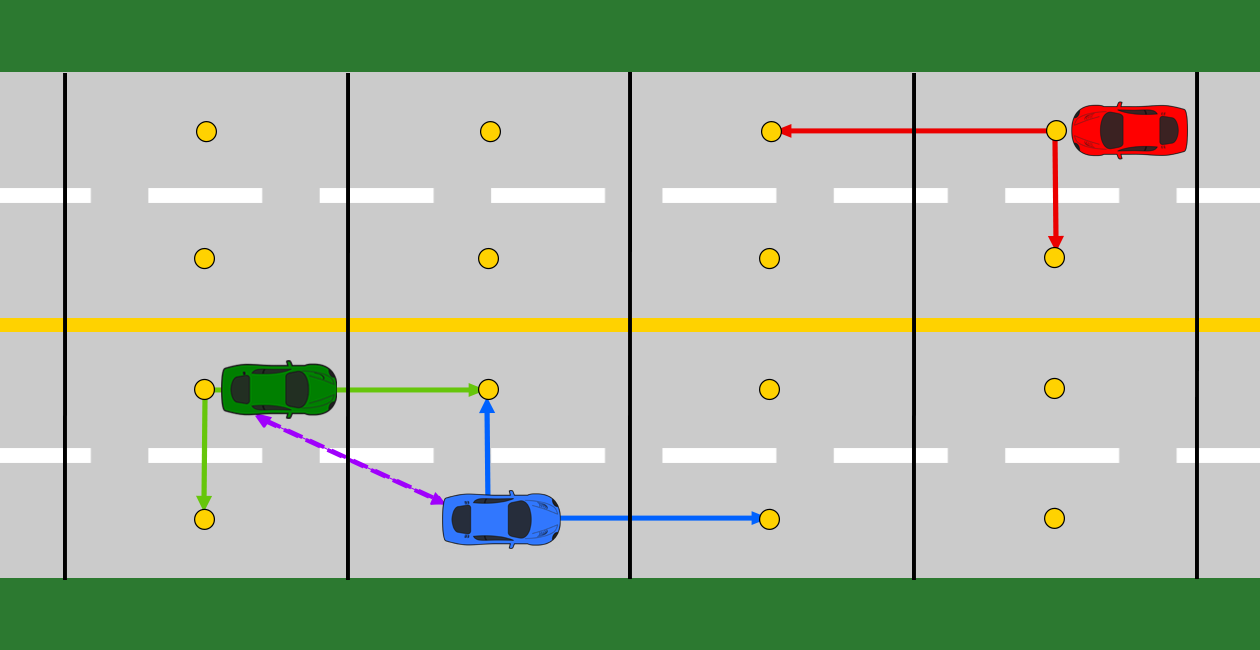}
    \vspace{-15pt}
    \caption{\textbf{Map discretization and interaction edges.} \textmd{We model the vehicle AD problem as an interaction graph with vehicle and lane nodes. A continuous map of the road is discretized into blocks. Directed lane edges between lane nodes encode permissible routes for vehicles. The red vehicle has a directed edge toward the lane nodes in front and to its left because the driver can legally continue forward or merge left. Conversely, the green vehicle has no edge connecting to a left lane node since it cannot cross the road divider. Vehicle edges, shown in purple, exist for vehicles that are close enough to interact with each other.}}
    \label{fig:lane_setup}
    \vspace{-15pt}
\end{figure}

\section{Related Works}
\label{sec:related}

\subsection{Exploiting Map Information}

The quality of information about an environment provided by High Definition maps (HD-maps) has dramatically increased and led to their ubiquitous use due to recent advancements in autonomous driving~\cite{mi2021hdmapgen, Yoon2022design}. 
Currently, most state-of-the-art methods for vehicle trajectory prediction, motion forecasting, and anomaly detection, do not make effective use of the rich information provided in these HD-maps, and only rely on modeling the interactions between vehicles on the road~\cite{chai2020multipath, lee2017desire, Shah2020LiRaNetET}.
Hence, these methods ignore vital information such as the plausible movement of vehicles in the environment, which can be paramount in identifying anomalies such as wrong-way driving. 

However, trajectory prediction methods such as those proposed by Deo \etal and Liang \etal do exploit the information in these HD-maps and significantly outperform their counterparts~\cite{LANEGCN2020, deo2021multimodal}. 
In proposing LaneGCN, Liang \etal encode different types of interactions between agents on the road with lane information extracted from maps.~\cite{LANEGCN2020}. 
They show that attention-based models can be used to encode interactions between vehicles and lanes, which are learned by constructing a graph representation of the road.
PGP, proposed by Deo \etal, further produces scene-compliant trajectories by sampling from a distribution of driving profiles conditioned on environment and vehicle interactions~\cite{deo2021multimodal}.
We corroborate the usefulness of these vehicle and lane attention-based representations and show that such embeddings do in fact provide meaningful insights in detecting highway vehicle anomalies in SABeR-VAE.


\subsection{Variational Autoencoders for Sequences}
Variational autoencoders (VAE) have been applied to sequential data combined with recurrent neural networks (RNN) in fields such as speech and image synthesis and autonomous driving~\cite{bl2012modeling, choi2022multi, gregor2015draw, chung2015vrnn, costa2021semi, liu2022learning, park2018multimodal}.
Liu \etal attempt to infer the traits of drivers from trajectories encoded in a variational latent space~\cite{liu2022learning}. 
However, only two classes of traits and a restricted set of defined trajectories were considered, while real drivers have a much wider range of behaviors on the road. 
Furthermore, they do not utilize map information in their learning process, which provide relevant context for traits.
Conditional VAE formulations have also been found to be able to generate trajectories with different driving styles, but fail to consistently produce feasible trajectories without necessary environment context~\cite{NIPS2015_8d55a249, ivanovic2020multimodal, salzmann2020trajectron++}.
Recurrent VAEs have also been applied to robot anomaly detection, but are limited by the simplicity of the single agent problem statement~\cite{park2018multimodal}. 
These sequential generative modeling approaches perform reasonably on their simple tasks, but fail to generate realistic samples from points in the latent space in more complex areas, due to the limitations of their RNN components~\cite{han2021fault, choi2022multi, chung2015vrnn}. 


To bridge the gap between complex human behaviors and the structured environment, and overcome the hurdles of the temporal propagation in simplistic RNNs, we propose the use of a lane-conditioned Koopman Operator to model the temporal relations in the latent space.
We were specifically inspired to use the Koopman operator to propagate the latent space due to its capability to model the dynamics of complex, non-linear data, including fluid dynamics, battery properties, and control tasks~\cite{PhysRevFluids.2.124402, kaushik_sak, balakrishnan2020deep, ijcai2019p440}.



\subsection{Anomaly Detection} 
Anomaly detection is well studied in diverse research areas and application domains~\cite{pang2021deep,chandola2009anomaly}. 
In robotics and automated vehicles, AD has been used to detect abnormal patterns such as robot failures~\cite{park2016multimodal,ji2020multi} and dangerous driving scenarios~\cite{yao2019unsupervised,MAAD2021}.

Park \etal propose a long short-term memory based variational autoencoder (LSTM-VAE) to reconstruct the expected distribution of robot sensor signals. 
A reconstruction-based anomaly score is then used for anomaly detection~\cite{park2018multimodal}. 
Furthermore, Ji \etal adopt an attention mechanism to fuse multi-sensor signals for robust anomaly detection in uncertain environments~\cite{ji2022proactive}. 
While these approaches focus on AD for single agent problem statements, our highway scenarios consist of complex multi-agent social interactions among vehicles, and need to be modeled as such.

In the domain of traffic anomaly detection using multi-agent trajectories, the most similar work to ours is the spatio-temporal graph auto-encoder (STGAE) proposed with the MAAD dataset~\cite{MAAD2021}. 
The architecture follows an encoder-decoder structure to reconstruct vehicle trajectories, where vehicle interactions and motions are considered using spatial graph convolution and temporal convolution layers, respectively. 
The method has been shown to be effective by modeling interactions among vehicles to detect anomalous maneuvers in traffic.
However, such a network ignores the constraints imposed by road structures on vehicle trajectories and the variability of human driver behaviors. 
In this work, we explicitly model both vehicle-to-vehicle interactions and lane-to-vehicle interactions to boost performance, and use an interpretable variational architecture to learn a continuous distribution over behaviors.

\section{Methodology}
\label{sec:methods}

In this section, we first introduce our problem formulation of anomaly detection from vehicle trajectories, and then explain our proposed SABeR-VAE framework.

\subsection{Problem Formulation}

Suppose $n_t\in[1, N]$ vehicles are on a road segment at any time $t$, and each vehicle takes an acceleration and steering action every timestep according to unknown policies. 
Let $c_t^{(i)} = \left(x_t^{(i)}, y_t^{(i)}\right)$ be the $2D$ coordinates of the $i^{\text{th}}$ vehicle at time $t$, where $i\in[1, ..., n_t]$. 
Each vehicle also has a set of corresponding permissible lane positions in front, to the left, and to the right of the vehicle, provided in the form of a discretized map representation shown in Fig.~\ref{fig:lane_setup}.
At every timestep, each vehicle's position within the map is used to identify their corresponding front, left, and right lane nodes. 
We define a tuple $l_t^{(i)}=\left(\text{front}, \text{left}, \text{right}\right)_t^{(i)}$ of three $2D$ coordinates containing the lane information for vehicle $i$ at time $t$. 
The discretization step only impacts $l_t^{(i)}$.
Altogether, the observed information of each vehicle at any time is the relative displacement of coordinates $o_t^{(i)} = \left(c_t^{(i)}-c_{t-1}^{(i)}, l_t^{(i)}-c_t^{(i)}\right) = \left(X_t^{(i)}, L_t^{(i)}\right)$.
A trajectory of length $T$ for any vehicle is represented as $\left(o_0^{(i)}, o_1^{(i)}, ..., o_{T-1}^{(i)}\right)$.
We assume that any vehicle $A$ that is within a distance $d$ to another vehicle $B$ at time $t$ can accurately detect and track the relative coordinates $c_t^{(B)}-c_t^{(A)} = R_t^{(A,B)}$.
The purple arrow between the green and blue vehicle in Fig.~\ref{fig:lane_setup} represents this vehicle interaction type.
For the $i$-th car, the number of observable cars is $m^i_t\in[0, n_t-1]$.
Given all vehicle trajectories in a scene, our goal is to provide an anomaly score $\text{AS}_t\in\mathbb{R}_{\geq0}$ for each time $t$.

\subsection{Architecture}

\begin{figure*}[tbp]
    \centering
    \includegraphics[width=0.95\textwidth]{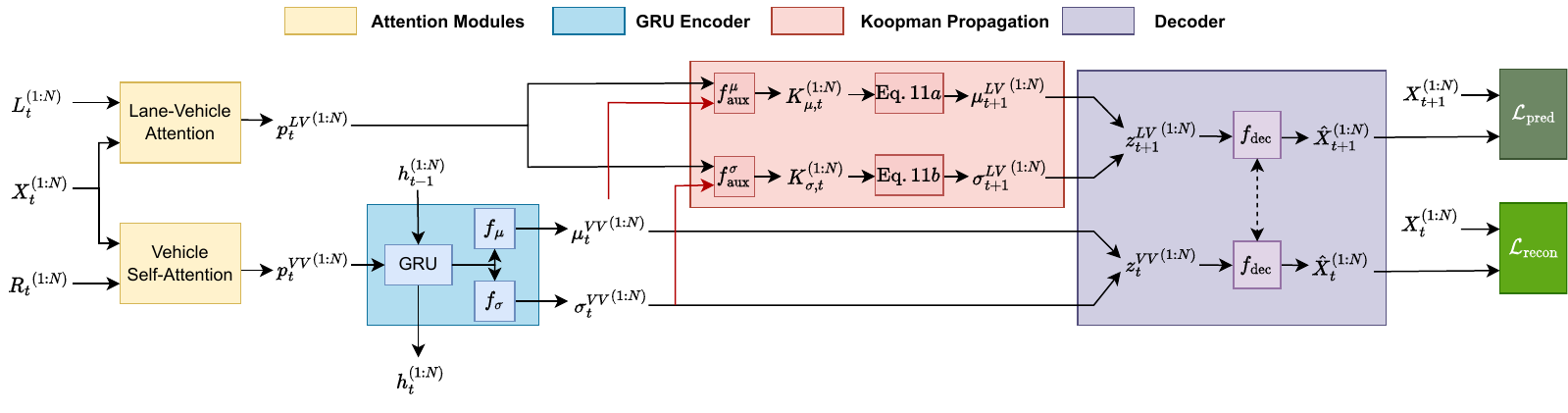}
    \vspace{-10pt}
    \caption{\textbf{SABeR-VAE architecture.} \textmd{The SABeR-VAE architecture attempts to predict the one-step future states of vehicles conditioned on current vehicle positions and structural lane information. Vehicle interactions are modeled by the self-attention module while permissible routes are encoded by the lane-vehicle attention module. A GRU encoder processes the self-attention embeddings through time to produce a latent distribution. Then the Koopman operator conditioned on the lane embeddings propagates the latent distributions forward, which finally get decoded to predict next states. The $f_{\text{dec}}$ network shares parameters for reconstruction and prediction.}}
    \label{fig:net}
    \vspace{-10pt}
\end{figure*}

Figure~\ref{fig:net} contains the complete architecture diagram of SABeR-VAE, which we discuss in this section.

\subsubsection{\textbf{Vehicle-Vehicle Self-Attention Network}}

Our goal is to learn a representation of spatial interactions among vehicles. 
Rather than using convolutional methods like those in prior works~\cite{MAAD2021, LANEGCN2020}, we encode the positions of vehicles on the road at each time with scaled dot-product multi-head self-attention, which allows each head to learn different features of the data~\cite{attention}. 
We embed the displacement of each car $X_t$ with a multi-layer perceptron (MLP) $f_Q^{VV}$ to obtain queries $Q_t^{VV}$:
\begin{equation}
    f_Q^{VV}(X_t) = Q_t^{VV} \in \mathbb{R}^{1\times D},\label{eq:one} 
\end{equation}
where $D$ is the attention size.
Let $\mathbf{R}_t = \left[R_t^{(i, 1)}, ..., R_t^{(i, m)}\right]^\intercal$ be the displacements of all neighboring cars for the $i$-th car. We use two other MLPs, $f_K^{VV}$ and $f_V^{VV}$, to embed $\mathbf{R}_t$ to obtain keys $K_t^{VV}$ and values $V_t^{VV}$ respectively:
\begin{align}
\begin{split}
f_K^{VV}(\mathbf{R}_t) &= K_t^{VV} \in \mathbb{R}^{m\times D} \\
f_V^{VV}(\mathbf{R}_t) &= V_t^{VV} \in \mathbb{R}^{m\times D}
\end{split}
\end{align}

The final encoding of each vehicle position from this self-attention layer for time $t$ is calculated as:
\begin{equation} \label{eq:3}
\text{softmax}\left(\frac{Q_t^{VV} {\left(K_t^{VV}\right)}^\intercal}{\sqrt{D}}\right)V_t^{VV} = p_t^{VV} \in \mathbb{R}^{1\times D}
\end{equation}


Nonexistent or unobserved vehicles further than a distance $d$ cannot be allowed to contribute to the attention score of other vehicles. Thus, we use a mask to set the score contributed from unobserved vehicles to $-\infty$. 

\subsubsection{\textbf{Lane-Vehicle Attention Network}}


We use available map and lane information in a separate lane-vehicle attention layer to model legal maneuvers in structured environments.
Similar to vehicle-vehicle attention, the query $Q^{LV}_t$ is an embedding of $X_t$.
Lane information of each vehicle $L_t$
is used to produce keys $K^{LV}_t$ and values $V^{LV}_t$:
\begin{align}
\begin{split}
f^{LV}_Q(X_t) &= Q^{LV}_t \in \mathbb{R}^{1\times D} \\
f^{LV}_K(L_t) &= K^{LV}_t \in \mathbb{R}^{3\times D} \\
f^{LV}_V(L_t) &= V^{LV}_t \in \mathbb{R}^{3\times D}
\end{split}
\end{align}



The lane-conditioned vehicle embeddings are calculated as:
\begin{equation} \label{eq:5}
\text{softmax}\left(\frac{Q^{LV}_t\left({K^{LV}_t}\right)^\intercal}{\sqrt{D}}\right)V^{LV}_t = p^{LV}_t \in \mathbb{R}^{1\times D}
\end{equation}

Note that all three lane nodes may not always be permissible to a vehicle. For example, a car in the left-most lane of a road is unable to legally turn left. As such, we mask out impermissible lane nodes like in the self-attention layer.
\subsubsection{\textbf{Recurrent Encoder}}

A gated recurrent unit (GRU) network encodes the sequence of self-attention features for each vehicle $\left(p_0^{VV}, p_1^{VV}, ..., p_{T-1}^{VV}\right)$ into a sequence of Gaussian distributions in the latent space with temporal correlation. Thus, the latent space captures the stochastic nature of human behaviors. 
Specifically, after embedding the vehicle-vehicle attention feature with a network $f_e$, we pass the embedding through the GRU to get the hidden state of each vehicle for the current timestep:
\begin{equation} \label{eq:6}
h_t = \text{GRU}\left( h_{t-1}, f_e\left(p_t^{VV}\right) \right)
\end{equation}

Mean and variance neural networks $f_\mu$ and $f_\sigma$ produce parameters for a latent normal distribution of dimension $j$ conditioned on a vehicle's hidden state at any time:
\begin{equation} \label{eq:7}
\mu_t^{VV} = f_\mu\left(h_t\right), \quad \sigma_t^{VV} = f_\sigma\left(h_t\right).
\end{equation}


\subsubsection{\textbf{Latent Propagation with Koopman Operator}}

While the GRU encoder encodes vehicle behaviors into the latent space solely conditioned on past and current vehicle interactions, we need a method to propagate the latent distributions in time to predict the future states of vehicles. 
To this end, we learn a stochastic Koopman operator conditioned on the lane-vehicle embeddings to perform this task, like Balakrishnan and Upadhyay~\cite{kaushik_sak}. 
The Koopman operator is responsible for temporal reasoning (modeling vehicle state dynamics), while the preceding attention modules take charge of spatial reasoning.

In Koopman operator theory, a discrete time system evolves according to potentially nonlinear dynamics $x_{t+1}=F(x_t)$. However, a function $g$ maps the state $x_t$ into a space where dynamics evolve linearly with the Koopman operator $\mathcal{K}$~\cite{kaushik_sak}:
\begin{equation}
    \mathcal{K}g\left(x_t\right) = g\left(F\left(x_t\right)\right) = g\left(x_{t+1}\right)
\end{equation}

Similarly, the inverse of function $g$ translates an observable of $x$ back into the original dynamics space~\cite{kaushik_sak}:
\begin{equation}\label{eq:g_minus_one}
    g^{-1}\left(\mathcal{K}g\left(x_t\right)\right) = x_{t+1}
\end{equation}

In our case, function $g$ is represented by the GRU encoder and neural networks $f_\mu$ and $f_\sigma$, which altogether, produce a latent distribution $\mathcal{N}\left(\mu_t, \sigma_t\right)$ conditioned on inter-vehicle embeddings $p_t^{VV}$. 

Like the Stochastic Adversarial Koopman (SAK) model~\cite{kaushik_sak}, we use auxiliary neural networks $f_{\text{aux}}^\mu$ and $f_{\text{aux}}^\sigma$ to predict tridiagonal Koopman matrices $K_{\mu, t}$ and $K_{\sigma, t}$, rather than solving for their closed form solution. 
The outputs of $f_{\text{aux}}^\mu$ and $f_{\text{aux}}^\sigma$ are conditioned on the current latent distributions $\mathcal{N}\left(\mu_t^{VV}, \sigma_t^{VV}\right)$ and the lane features $p_t^{LV}$, so that the Koopman operators capture legal route maneuvers in the latent space propagation:
\begin{align}
\begin{split}
K_{\mu, t} &= f_{\text{aux}}^\mu\left(\mu_t^{VV}, p_t^{LV}\right) \\
K_{\sigma, t} &= f_{\text{aux}}^\sigma\left(\sigma_t^{VV}, p_t^{LV}\right)
\end{split}
\end{align}

The predicted Koopman matrices are applied to the inter-vehicle distributions to linearly propagate the mean and variance of the latent distributions forward in time:
\begin{subequations}
\label{eq:8}
\begin{align}
\mu_{t+1}^{LV} &= K_{\mu, t}\:\mu_{t}^{VV} + \mu_{t}^{VV} \\
\sigma_{t+1}^{LV} &= K_{\sigma, t}\:\sigma_{t}^{VV} + \sigma_{t}^{VV}
\end{align}
\end{subequations}

Intuitively, we can interpret the GRU encoder as predicting a distribution of vehicle behaviors from their current trajectories, and the Koopman operator propagates to a one-step future distribution of behaviors based on lane information. 


\subsubsection{\textbf{The Decoder Network}}

At this point, we have two sets of distributions in the $j$-dimensional latent space for the current states and future predictions of vehicles at each time: $\mathcal{N}\left(\mu_t^{VV}, \sigma_t^{VV}\right)$ and $\mathcal{N}\left({\mu}_{t+1}^{LV}, {\sigma}_{t+1}^{LV}\right)$. We utilize the reparameterization trick to sample a point from each of the distributions:
\begin{equation}
\begin{alignedat}{2}
\epsilon_t^{VV} &\sim \mathcal{N}\left(0, 1\right) &\quad z_t^{VV} &= \mu_{t}^{VV} + \epsilon_t^{VV} \sigma_{t}^{VV} \\
\epsilon_{t+1}^{LV} &\sim \mathcal{N}\left(0, 1\right) &\quad z_{t+1}^{LV} &= {\mu}_{t+1}^{LV} + \epsilon_{t+1}^{LV} \sigma_{t+1}^{LV}
\end{alignedat}
\end{equation}

A multi-layer perceptron $f_{\text{dec}}$ is used as a decoder network, similar to $g^{-1}$ in Eq.~\ref{eq:g_minus_one}, to predict a vehicle coordinate change from the sampled latent points:
\begin{equation} \label{eq:9}
\hat{X}_t = f_{\text{dec}}\left(z_t^{VV}\right), \qquad \hat{X}_{t+1} = f_{\text{dec}}\left(z_{t+1}^{LV}\right).
\end{equation}

\subsection{Training and Evaluation}
\subsubsection{\textbf{End-to-End Training}}

To fairly compare our method with prior convolutional approaches, we utilize a similar sliding window training approach performed by Wiederer \etal~\cite{MAAD2021}. Specifically, whole trajectories of length $T$ are divided into small overlapping segments, or windows, of constant length $T'$.

In our training objective, we minimize the current reconstruction loss and one-step future prediction loss of the model by splitting our input ground truth trajectories into current states $X^- = X_{0:T'-2}$ and one-step future states $X^+ = X_{1:T'-1}$. We also regularize the current distributions $\mathcal{N}\left(\mu_t^{VV}, \sigma_t^{VV}\right)$ and propagated distributions $\mathcal{N}\left({\mu}_{t+1}^{LV}, {\sigma}_{t+1}^{LV}\right)$ to follow a standard normal distribution. Let $D_{\text{KL}}(\mu, \sigma)$ be the KL divergence between any Gaussian distribution $\mathcal{N}(\mu, \sigma)$ and the standard normal distribution $\mathcal{N}\left(0,1\right)$. Then the regularized prediction and reconstruction losses are:
\begin{equation}
\begin{alignedat}{2}
\mathcal{L}_{\text{pred}} = \beta_1 \cdot D_{\text{KL}}\left(\mu^{LV},\sigma^{LV}\right) + \lVert X^+ - \hat{X}^+ \rVert_2 \\
\mathcal{L}_{\text{recon}} = \beta_2 \cdot D_{\text{KL}}\left(\mu^{VV},\sigma^{VV}\right) + \lVert X^- - \hat{X}^- \rVert_2,
\end{alignedat}
\end{equation}
where $\beta_1$ and $\beta_2$ are tunable weights applied to the regularization of the latent distributions similar to beta-VAE~\cite{Higgins2017betaVAELB}.



The overall objective we optimize is:
\begin{equation} \label{eq:L}
\mathcal{L} = \mathcal{L}_{\text{pred}} + \mathcal{L}_{\text{recon}}
\end{equation}
We again mask out coordinates of unobserved vehicles so they do not contribute to the loss. 

While SAK~\cite{kaushik_sak} applies maximum mean discrepancy (MMD) to synchronize the current and propagated distributions of their Koopman model to any general distribution, we explicitly encourage the latent space distributions to follow the standard gaussian. We leave experimentation of various Koopman synchronization methods for the anomaly detection task as a future direction of research. 

\subsubsection{\textbf{Anomaly Detection Evaluation}}

At test time, we follow the same sliding window practice as performed in training. First, we calculate the one-step future prediction loss $\mathcal{L}_{\text{pred},{t+1}} = \lVert X_{t+1}^{(i)} - \hat{X}_{t+1}^{(i)} \rVert_2$ for every vehicle at each timestep, within every window of a complete trajectory. 

Then, we average the prediction loss of overlapping timesteps among all windows in the sequence, for each vehicle separately. 
Suppose $\mathcal{W}^{(t,i)}$ is the set of all windows in the complete trajectory containing time $t$ where vehicle $i$ is observed.
The averaged prediction error for car $i$ at $t$ is:
\begin{equation}
    \bar{\mathcal{L}}_{\text{pred}, t}^{(i)} = \frac{\sum_{w\in\mathcal{W}^{(t,i)}}{\mathcal{L}_{\text{pred}, {w_t^{(i)}}}}}{\lvert\mathcal{W}^{(t,i)}\rvert},
\end{equation}
where $\mathcal{L}_{\text{pred}, {w_t^{(i)}}}$ is the prediction error of time $t$ for vehicle $i$ in window $w$ of the set $\mathcal{W}^{(t,i)}$.

Finally, we choose the anomaly score AS to be the maximum averaged prediction loss over all vehicles at a given timestep $t$:
\begin{equation}
    \text{AS}_t = \max_{i=1, ..., n_t}{\bar{\mathcal{L}}_{\text{pred}, t}^{(i)}}
\end{equation}


\begin{figure}[]
    \centering
    \includegraphics[width=\linewidth]{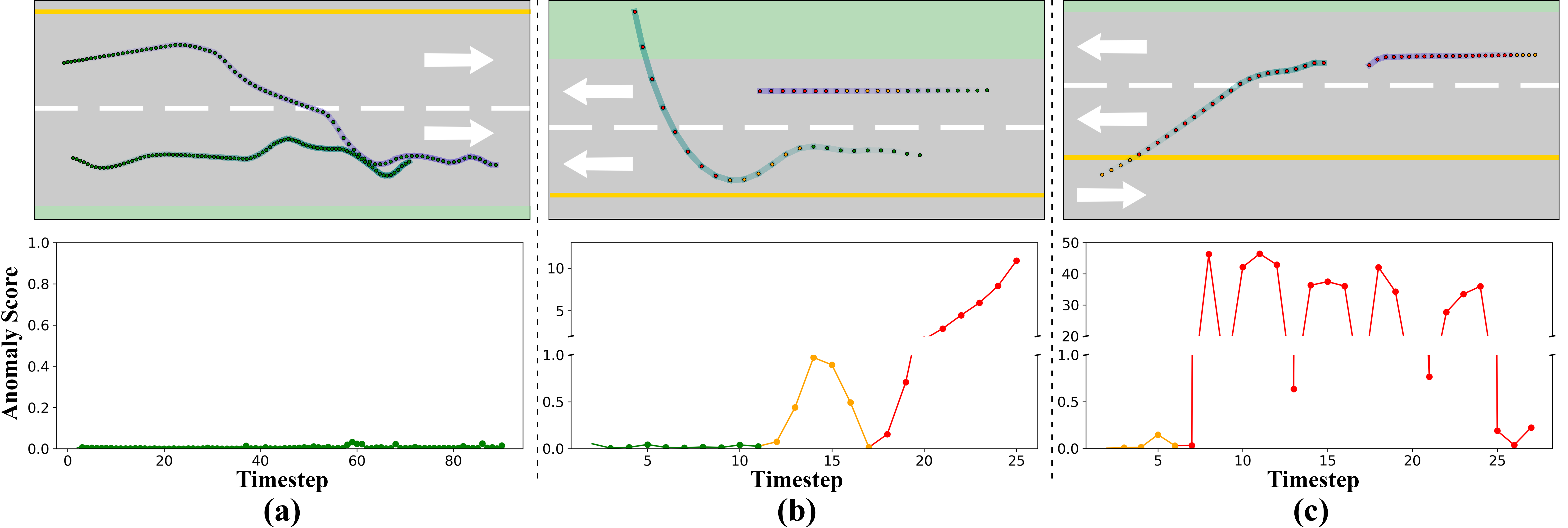}
    \caption{\textbf{Trajectories and SABeR-VAE anomaly scores.} \textmd{\textbf{(top row)} Examples of a normal overtaking (a,) abnormal off-road driving (b,) and wrong-way driving (c) scenarios in the MAAD dataset. White arrows point toward direction of normal traffic flow. \textbf{(bottom row)} Predicted anomaly score curves for each scenario above. Colors of lines within the curves show the ground-truth labels of normal (green,) ignored (yellow,) \& abnormal (red) timesteps.}}
    \label{fig:traj}
    \vspace{-10pt}
\end{figure}

\section{Experimental setup and results}
\label{sec:results}

\begin{table*}[ht]
\centering
\caption{\textbf{Accuracy results of baselines, ablations, and SABeR methods over ten runs.}}
\label{tab:accuracy}
\begin{threeparttable}
\begin{tabular}{lccccc}
\toprule
Method                     & Detection Type      & AUROC $\uparrow$          & AUPR-Abnormal $\uparrow$  & AUPR-Normal $\uparrow$    & FPR @ 95\%-TPR $\downarrow$ \\ 
\midrule
CVM                          & Reconstruction Loss & $83.1 \pm 0.0$            & $54.5 \pm 0.0$            & $96.0 \pm 0.0$            & $74.6 \pm 0.0$              \\
RAE-Recon\tnote{*}           & Reconstruction Loss & $56.2 \pm 0.7$            & $16.9 \pm 1.0$            & $89.5 \pm 0.1$            & $84.6 \pm 0.3$              \\
STGAE\tnote{*}               & Reconstruction Loss & $74.8 \pm 5.1$            & $37.8 \pm 7.2$            & $94.1 \pm 1.3$            & $77.8 \pm 9.8$              \\
STGAE-KDE\tnote{*}           & One Class           & $86.3 \pm 1.7$            & $55.2 \pm 7.7$            & \textbf{97.2 $\pm$ 0.5}   & \textbf{50.0 $\pm$ 7.9}     \\ 
\midrule
RAE-Pred                     & Prediction Loss     & $72.5 \pm 15.3$           & $43.5 \pm 17.4$           & $92.9 \pm 4.4$            & $75.8 \pm 10.6$             \\
VV-RAE                       & Prediction Loss     & $54.2 \pm 4.9$            & $14.8 \pm 0.9$            & $89.5 \pm 2.4$            & $77.1 \pm 7.3$              \\ 
Att-LSTM-VAE                 & Prediction Loss     & $85.8 \pm 0.7$            & $64.9 \pm 0.9$            & $96.5 \pm 0.3$            & $66.6 \pm 5.9$              \\
\midrule
SABeR-AE                     & Prediction Loss     & \textbf{87.2 $\pm$ 0.4}   & \textbf{69.0 $\pm$ 0.5}   & $96.9 \pm 0.2$            & $64.1 \pm 5.0$              \\ 
SABeR-VAE                    & Prediction Loss     & $87.0 \pm 1.5$            & $65.5 \pm 2.9$            & $96.9 \pm 0.5$            & $57.7 \pm 7.6$              \\ 
\bottomrule
\end{tabular}
\begin{tablenotes}
\item[*] These results are presented in~\cite{MAAD2021}.
\end{tablenotes}
\end{threeparttable}
\vspace{-10pt}
\end{table*}

In this section, we first describe the MAAD dataset on which we performed experiments and detail baselines and ablations. We also present our quantitative results and latent space interpretations.

\subsection{MAAD Dataset and Augmentation}

The MAAD dataset~\cite{MAAD2021} consists of $2D$ trajectories of two vehicles on a straight two-lane highway with a divider separating the two possible directions, as visualized in the top row of Fig.~\ref{fig:traj}.
There are 80 training and 66 test-split trajectories ranging from a length of 25 to 127 timesteps.
To compare fairly with baselines, these dynamic length trajectories are subsampled to produce approximately 6.3K training and 3.1K testing windows of constant length $T'=15$. 
As the original dataset sequences did not come with map or lane details, we augmented the data to include this information. 
Specifically, we discretized the highway in the $x$-coordinate direction into blocks of length five meters as shown in Fig.~\ref{fig:lane_setup}, and stored the $2D$ coordinates of the front, left, and right blocks for each vehicle at every timestep in all trajectories. 
We chose a discretization factor of five meters because vehicles traveled on average five meters or less every timestep.
All the training sequences consist of normal vehicle behaviors like driving side-by-side, overtaking, following, and driving in opposite directions. 
In contrast, the test-split contains both normal and $11$ anomalous behavior classes like aggressive overtaking, pushing aside, tailgating, off-road, and wrong-way driving.

\subsection{Baseline Methods}

We compare against baselines implemented by Wiederer \etal that depend on reconstruction loss rather than future prediction error~\cite{MAAD2021}. 
\textbf{(1)} The Constant Velocity Model (\textbf{CVM}) is a standard baseline that predicts the next states of vehicles assuming each vehicle travels at the same velocity as the last timestep, without modeling any inter-vehicle relations. 
\textbf{(2)} Recurrent Autoencoder (\textbf{RAE-Recon}) uses an LSTM network to encode and decode a sequence of coordinates from an unregularized latent space, attempting to minimize reconstruction loss. 
\textbf{(3)} Spatio-Temporal Graph Autoencoder (\textbf{STGAE}) is a convolutional method that models inter-vehicle behaviors, and outputs parameters for a bi-variate distribution describing the estimated state of the reconstructed pose of vehicles, and is trained to maximize the log-likelihood of the estimated probability distribution. 
Finally, \textbf{(4)} the \textbf{STGAE-KDE} baseline fits a Kernel Density Estimator (KDE) to the unregularized latent space of a trained STGAE model to predict the one-class probability of a set of points originating from a normal behavior window. 
Unlike the STGAE-KDE, our SABeR-VAE does not require an expensive KDE fitting procedure since our anomaly score solely relies on prediction error, and we still model inter-vehicle relations unlike CVM and RAE-Recon.

We additionally train ablation models with future prediction loss to identify the impact of different components in our method. 
We train \textbf{(5)} an unregularized Recurrent Autoencoder (\textbf{RAE-Pred},) using a standard deterministic MLP to propagate latent points forward in time, without explicitly modeling any inter-vehicle behaviors, like the RAE-Recon. 
\textbf{(6)} A Recurrent Autoencoder with a vehicle-vehicle Self-Attention module (\textbf{VV-RAE}) minimizes prediction error while modeling inter-vehicle relations. 
We also train \textbf{(7)} a deterministic variant of SABeR-VAE without a regularized latent space, \textbf{SABeR-AE}. SABeR-AE utilizes both vehicle self-attention and lane-vehicle attention like SABeR-VAE, but encodes trajectories into an unregularized (uninterpretable) latent space.
\textbf{(8)} To test the effectiveness of the Koopman operator in SABeR-VAE, we train an ablation model \textbf{(Att-LSTM-VAE)} that replaces the Koopman propagation module with a recurrent decoder like Park \etal~\cite{park2018multimodal}.

\subsection{Quantitative Evaluation Metrics}


We quantitatively evaluate the effectiveness of models on the MAAD dataset using four metrics. \textbf{(1)} Area Under Receiver-Operating Characteristic curve (\textbf{AUROC}) is calculated by plotting the False-Positive Rate (FPR) and True-Positive Rate (TPR) of a model over several decision thresholds, and computing the area under the curve. A model with greater AUROC performs better, and a perfect classifier has an AUROC of $100\%$. Though, AUROC is skewed in datasets where there are very few positive labels, like in the field of outlier identification. As such, FPR may be misleadingly low, producing an optimistic AUROC value. We compute \textbf{(2)} the Area Under Precision-Recall Curve (AUPR) with the anomalous points being the positive class (\textbf{AUPR-Abnormal}) and \textbf{(3)} with normal points being positive (\textbf{AUPR-Normal}). The AUPR metric adjusts for skewed dataset distributions, and we evaluate model effectiveness of classifying anomalies, and not mis-classifying normal points with AUPR-Abnormal and AUPR-Normal respectively. Finally, we use \textbf{(4)} \textbf{FPR @ 95\%-TPR} to check the rate of mis-labeling normal points when TPR is high.



\subsection{Accuracy Results}

\begin{figure}
    \centering
    \vspace{-15pt}
    \includegraphics[width=0.9\columnwidth]{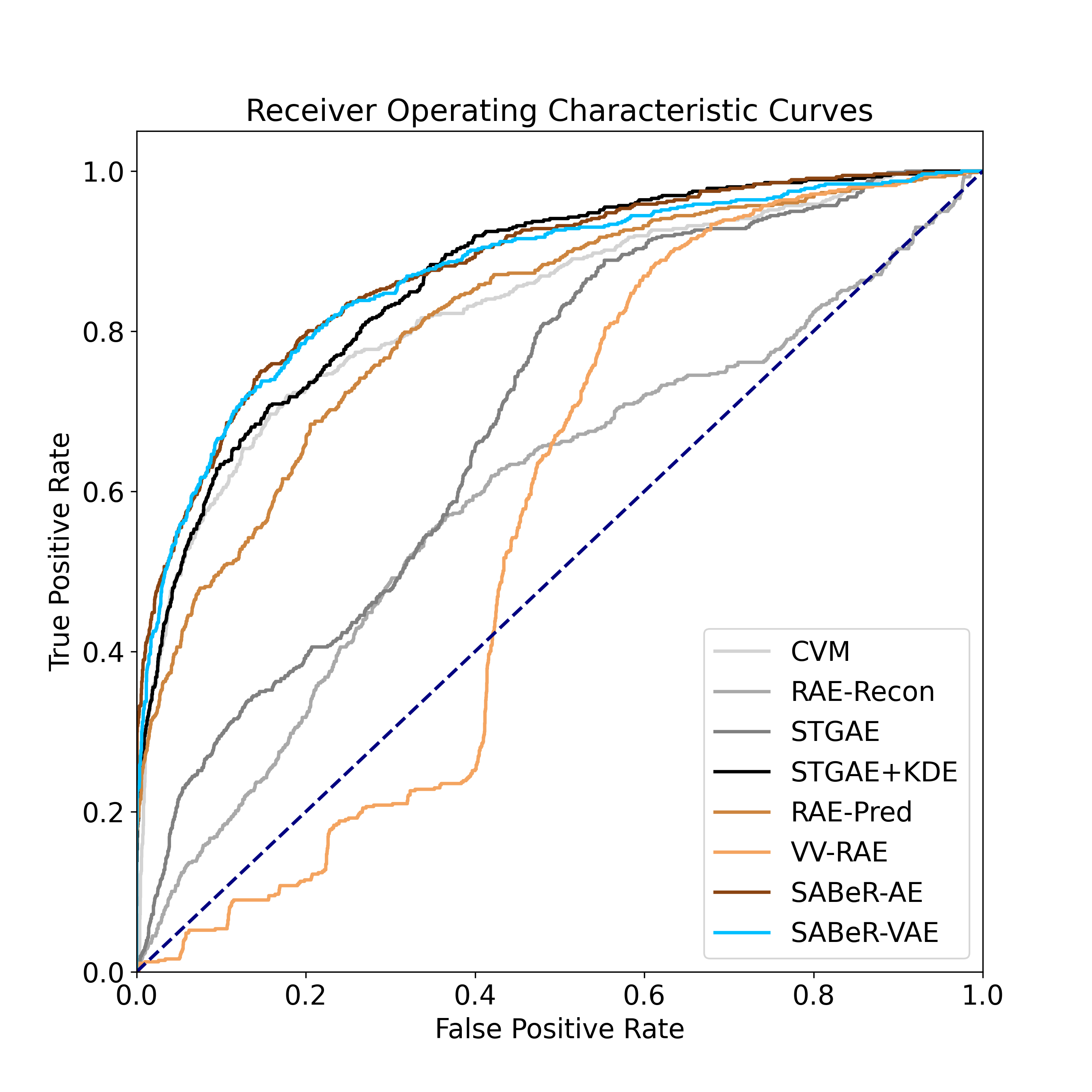}
    \vspace{-15pt}
    \caption{\textbf{ROC curves of tested methods.}}
    \label{fig:roc}
    \vspace{-10pt}
\end{figure}

Every model was trained for $500$ epochs on the training split with a Tesla V100 GPU~\cite{HAL}. Like Wiederer \etal, we calculate metrics for each hyperparameter choice on a $20\%$ validation split of the whole test data, and choose to evaluate the best set of hyperparameters for each respective method on the complete test split~\cite{MAAD2021}. More details on training and hyperparameter choices are provided in Appendix~\ref{sec:app_train}.
Table~\ref{tab:accuracy} holds accuracy results of baselines, ablations, and the SABeR methods on the test split of the MAAD dataset. Each method, besides CVM, was trained ten separate times with the same hyperparameters, and we report the average and standard deviation of each methods' results over the ten runs. Figure~\ref{fig:roc} plots the ROC curves for each method.

Amongst baselines, the simple CVM model already performs well as an anomaly detector since its AUROC is only $3\%$ less than that of the STGAE-KDE method. 
CVM also has no variation of results since it is a deterministic model that is not trained.
The LSTM-based RAE-Recon model is unable to effectively distinguish between anomalies and normal scenarios using reconstruction loss, since it does not model vehicle or lane information. 
While recurrent models encode current timestep features based solely on previous timesteps, temporal convolution methods extract information from the whole trajectory, which helps to predict a more accurate reconstruction.
Thus, the convolutional STGAE method drastically improves AUROC and AUPR scores over RAE-Recon.

\begin{table}[]
\centering
\caption{\textbf{AUROC ($\uparrow$) of methods by anomaly type.}}
\label{tab:anom_accuracy}
\resizebox{\linewidth}{!}{%
\begin{threeparttable}
\begin{tabular}{lcccc}
\toprule
Anomaly Type & CVM   & STGAE-KDE\tnote{*} & SABeR-AE & SABeR-VAE \\ 
\midrule
Reeving                          & \textbf{96.7} & 94.6     & 87.9    & 89.5     \\
Pushing Aside                    & \textbf{91.5} & 90.4     & 88.6    & 91.3     \\
Right Spreading                  & 87.7 & \textbf{96.2}     & 86.7    & 95.2     \\
Left Spreading                   & 90.9 & \textbf{96.6}     & 96.2    & 95.9     \\
Off-Road                         & 88.7 & 98.2     & 98.2    & \textbf{99.7}     \\
Skidding                         & 96.9 & 99.7     & $\sim$\textbf{100.0}    & 99.8     \\
Wrong-way                        & 63.2 & 73.2     & $\sim$\textbf{100.0}    & 99.3     \\ 
\bottomrule
\end{tabular}
\begin{tablenotes}
\item[*] These results are presented in~\cite{MAAD2021}.
\end{tablenotes}
\end{threeparttable}
}
\vspace{-15pt}
\end{table}

However, once we incorporate a latent propagation network and predict future timesteps, the RAE-Pred ablation increases AUROC over RAE-Recon by $29\%$ and even outperforms STGAE in the AUPR-Abnormal metric, without even modeling inter-vehicle behaviors. 
This result hints to the idea that recurrent networks learn to model normal behaviors more accurately with future prediction error, than reconstruction error of observed timesteps, which assists in AD performance.
Furthermore, recurrent methods are capable of reaching the same performance as convolutional methods, while relying only on past data points.
Still, RAE-Pred is shown to be unstable as it produces a high variance in results over the ten trained models. This variance was caused by two of the ten runs achieving only $45\%$ AUROC.
STGAE also has the highest variance in results among baselines since it is a stochastic method reconstructing a distribution over states, rather than the deterministic CVM and RAE-Recon approaches, but is more stable than RAE-Pred.

We see that adding a vehicle-vehicle self-attention layer in VV-RAE model actually hinders performance, and gives results similar to RAE-Recon.
Effectively, the vehicle-vehicle self-attention layer did not learn useful features for the future prediction task, and confused the model generations. 
This outcome could be a result of a low complexity neural network or a potentially poor choice for masking distance $d$.

The one-class prediction model STGAE-KDE fits a KDE to the latent space of the STGAE to learn a distribution of normal latent behaviors. 
As such, this one-class classification approach improves detection rates and training stability over the STGAE such that it outperforms other baselines. However, the fitting process of a KDE to a large dimensional space is a computationally complex and constrictive part the method. With gaussian regularization of the latent space, our SABeR-VAE clusters similar behaviors together and learns a latent distribution without fitting a KDE, which we discuss in~\ref{sec:latent}.

Finally, SABeR-AE and SABeR-VAE incorporate a lane-vehicle attention module to capture the effect of the structure of the environment on normal behaviors. 
We see that SABeR-AE outperforms all methods in AUROC and AUPR-Abnormal with low variance, showcasing the importance of modeling environment structure in this field.
SABeR-VAE performs slightly better than STGAE-KDE in AUROC, and significantly increases the AUPR-Abnormal score by $18\%$.
However, the stochasticity of the SABeR-VAE method hinders its reproducibility, and AUROC scores ranged from $84\%$ to $89\%$ over the ten training runs.
SABeR-VAE further decreases the average FPR @ $95\%$-TPR of SABeR-AE by $10\%$.
STGAE-KDE and the two SABeR approaches have similar AUPR-Normal. 
SABeR-VAE also outperforms Att-LSTM-VAE, meaning a recurrent decoder is unnecessary when using the Koopman operator.

\begin{figure*}[tbp]
    \centering
    \includegraphics[width=\linewidth]{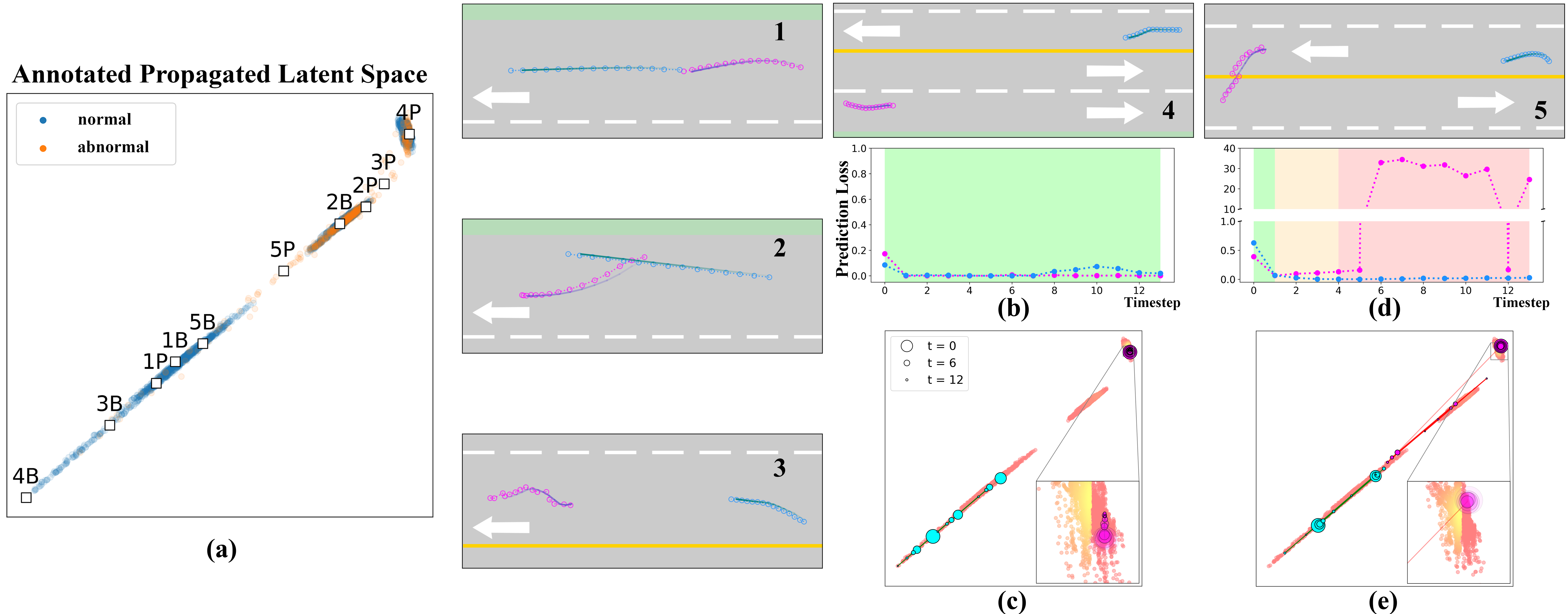}
    \caption{\textbf{Koopman propagated latent space and corresponding trajectories.} \textmd{\textbf{(a)} We encode every window trajectory in the test-split of the MAAD dataset and plot the $2D$ sampled latent positions of the final timestep of the windows. Blue points correspond to ground truth normal windows while orange are abnormal. \textbf{(1-5)} Five scenario windows are encoded into the latent space, and are explicitly annotated in (a.) Each of the five windows has two latent points for the pink and blue vehicles respectively. (e.g., annotations ``1B'' \& ``1P'' are the latent points of the blue and pink vehicles in road trajectory window 1.) Within the five trajectory windows, solid lines are the ground truth trajectory of the vehicle while open circles are predicted by SABeR-VAE. White arrows denote direction of traffic flow. \textbf{(b \& d)} The prediction error curves for trajectories 4 and 5 respectively. \textbf{(c \& e)} The trajectory of latent points for vehicle windows 4 and 5 respectively. A heat map of the original latent space is plotted in orange in the background. Blue and pink circles are the latent trajectories of the blue and pink car through time. The largest circles encode the initial timestep of the window, and they decrease in size as the window progresses.}}
    \label{fig:final_latent}
    \vspace{-10pt}
\end{figure*}

We present examples of SABeR-VAE scoring anomalous timesteps in Fig.~\ref{fig:traj}. There, a normal overtaking maneuver was scored very low during the whole trajectory, whereas going off-road or driving in the wrong direction were scored high. We can also see in Fig.~\ref{fig:traj}.b that timesteps where vehicles are acting normally prior to erratic behavior are still correctly scored low. Table~\ref{tab:anom_accuracy} holds AUROC by anomaly type for CVM, STGAE-KDE, and SABeR methods. SABeR-VAE improves wrong-way driving AD by $35\%$ over STGAE-KDE, while performing comparably in other metrics. 
The complete version of Table~\ref{tab:anom_accuracy} is provided in Appendix~\ref{sec:app_auroc}.

\vspace{-10pt}

\subsection{Latent Space Interpretation}\label{sec:latent}

SABeR-VAE is a variational model with a continuous latent space such that observations with similar learned characteristics are clustered closer together in the latent space. 
In Fig.~\ref{fig:final_latent}.a, we plot the test-split latent space of one of the SABeR-VAE models evaluated in Table~\ref{tab:accuracy}.
Points are clustered into three distinct regions in the latent space, which we will refer to as ``bottom,'' ``middle,'' and ``top'' clusters respectively. 
We see from sampled trajectories that the bottom and middle clusters encode vehicles that travel toward the $-x$ (left) direction in either of the top two lanes of the highway, while the top cluster encodes vehicles traveling to the right in the bottom two lanes. For example, points 1P, 2P, 3B, 4B, and 5B correspond to blue (B) and pink (P) cars that travel to the left in the top lanes. Similarly, the trajectory of the pink car driving to the right in window 4 is encoded to point 4P in the top cluster of the latent space. Vehicles that are also physically close and interacting with each other are encoded closely in the latent space, as shown with latent points corresponding to windows 1 and 2. The middle cluster embeds anomalous scenarios from the top lanes where vehicles are close enough to interact with each other, like window 2.

Furthermore, anomalous, non-interactive trajectories that are poorly predicted are encoded to the outskirts of the primary cluster distributions. For example, the pink cars in trajectories 3 and 5 are driving in the wrong direction. These trajectories have high prediction error as visualized by the little overlap between the predicted open circle positions and ground truth trajectories. As such, those poorly predicted points are encoded in the spaces between the bottom and middle, and middle and top clusters respectively. In contrast, trajectories 1 and 4 have low loss and are encoded to positions within the primary clusters. 
Thus the latent space has learned a correspondence between permissible lane routings and expected vehicle behavior. 

Finally, we visualize the transformation of the latent space over time within one trajectory window to show the interpretable impact of the learned Koopman operator. 
Figures~\ref{fig:final_latent}.c and~\ref{fig:final_latent}.e show the transformation of the latent space as time progresses in trajectory windows 4 and 5. 
We can see in Fig.~\ref{fig:final_latent}.c that the blue and pink latent trajectories stay in the bottom and top clusters respectively, since the vehicles follow the correct direction on the road throughout window 4. 
Conversely, we see in Fig.~\ref{fig:final_latent}.e that the pink latent trajectory begins in the top cluster since the pink vehicle in trajectory 5 is in one of the bottom two lanes on the road. 
But, at timestep 6, the pink car crosses the road divider into the wrong direction lane. Thus, we see a jump in the pink car's latent trajectory in Fig.~\ref{fig:final_latent}.e from the top cluster to the bottom and middle clusters that correspond to the top two lanes. 
At the same time, Fig.~\ref{fig:final_latent}.d has a spike in the prediction loss of the pink vehicle. 
For the remainder of the trajectory window, the pink car oscillates drastically in the latent space around the middle cluster, since the model expects the vehicle to be traveling to the left. 
Note, even though the pink vehicle in trajectory 5 is acting abnormally, this does not effect the latent trajectory of the blue vehicle in the same window, since the vehicles are not close enough to impact each other. 
Overall, the Koopman operator explicitly models this transition from normal to anomalous states in the latent space, in an interpretable manner.

\section{Conclusions and future work}
\label{sec:conclusion}

In this paper, we propose a novel framework for anomaly detection with an unsupervised recurrent VAE network conditioned on structured environment information and vehicle interactions. We show that modeling this structured information is imperative to having high accuracy over a wide range of anomaly types and study the interpretability of the architecture.
Future work includes using raw sensor data for detection and integrating with a vehicle controller.



\begin{acks}
We thank Julian Wiederer for providing access to the MAAD dataset and Kaushik Balakrishnan for insightful discussions. 
This work utilizes resources supported by the National Science Foundation’s Major Research Instrumentation program, grant $\#1725729$, as well as the University of Illinois at Urbana-Champaign.
\end{acks}



\bibliographystyle{ACM-Reference-Format} 
\balance
\bibliography{root}


\begin{thebibliography}{46}


\ifx \showCODEN    \undefined \def \showCODEN     #1{\unskip}     \fi
\ifx \showDOI      \undefined \def \showDOI       #1{#1}\fi
\ifx \showISBNx    \undefined \def \showISBNx     #1{\unskip}     \fi
\ifx \showISBNxiii \undefined \def \showISBNxiii  #1{\unskip}     \fi
\ifx \showISSN     \undefined \def \showISSN      #1{\unskip}     \fi
\ifx \showLCCN     \undefined \def \showLCCN      #1{\unskip}     \fi
\ifx \shownote     \undefined \def \shownote      #1{#1}          \fi
\ifx \showarticletitle \undefined \def \showarticletitle #1{#1}   \fi
\ifx \showURL      \undefined \def \showURL       {\relax}        \fi
\providecommand\bibfield[2]{#2}
\providecommand\bibinfo[2]{#2}
\providecommand\natexlab[1]{#1}
\providecommand\showeprint[2][]{arXiv:#2}

\bibitem[\protect\citeauthoryear{Arbabi and Mezi\ifmmode~\acute{c}\else
  \'{c}\fi{}}{Arbabi and Mezi\ifmmode~\acute{c}\else \'{c}\fi{}}{2017}]%
        {PhysRevFluids.2.124402}
\bibfield{author}{\bibinfo{person}{Hassan Arbabi} {and} \bibinfo{person}{Igor
  Mezi\ifmmode~\acute{c}\else \'{c}\fi{}}.} \bibinfo{year}{2017}\natexlab{}.
\newblock \showarticletitle{Study of dynamics in post-transient flows using
  Koopman mode decomposition}.
\newblock \bibinfo{journal}{\emph{Phys. Rev. Fluids}}  \bibinfo{volume}{2}
  (\bibinfo{date}{Dec} \bibinfo{year}{2017}), \bibinfo{pages}{124402}.
\newblock
Issue 12.
\urldef\tempurl%
\url{https://doi.org/10.1103/PhysRevFluids.2.124402}
\showDOI{\tempurl}


\bibitem[\protect\citeauthoryear{Azzalini, Castellini, Luperto, Farinelli, and
  Amigoni}{Azzalini et~al\mbox{.}}{2020}]%
        {azzalini2020hmms}
\bibfield{author}{\bibinfo{person}{Davide Azzalini}, \bibinfo{person}{Alberto
  Castellini}, \bibinfo{person}{Matteo Luperto}, \bibinfo{person}{Alessandro
  Farinelli}, {and} \bibinfo{person}{Francesco Amigoni}.}
  \bibinfo{year}{2020}\natexlab{}.
\newblock \showarticletitle{Hmms for anomaly detection in autonomous robots}.
  In \bibinfo{booktitle}{\emph{International Conference on Autonomous Agents
  and MultiAgent Systems}}. ACM, \bibinfo{pages}{105--113}.
\newblock


\bibitem[\protect\citeauthoryear{Balakrishnan and Upadhyay}{Balakrishnan and
  Upadhyay}{2020}]%
        {balakrishnan2020deep}
\bibfield{author}{\bibinfo{person}{Kaushik Balakrishnan} {and}
  \bibinfo{person}{Devesh Upadhyay}.} \bibinfo{year}{2020}\natexlab{}.
\newblock \showarticletitle{Deep adversarial koopman model for
  reaction-diffusion systems}.
\newblock \bibinfo{journal}{\emph{arXiv preprint arXiv:2006.05547}}
  (\bibinfo{year}{2020}).
\newblock


\bibitem[\protect\citeauthoryear{Balakrishnan and Upadhyay}{Balakrishnan and
  Upadhyay}{2021}]%
        {kaushik_sak}
\bibfield{author}{\bibinfo{person}{Kaushik Balakrishnan} {and}
  \bibinfo{person}{Devesh Upadhyay}.} \bibinfo{year}{2021}\natexlab{}.
\newblock \showarticletitle{Stochastic Adversarial Koopman Model for Dynamical
  Systems}.
\newblock \bibinfo{journal}{\emph{CoRR}}  \bibinfo{volume}{abs/2109.05095}
  (\bibinfo{year}{2021}).
\newblock
\showeprint[arXiv]{2109.05095}
\urldef\tempurl%
\url{https://arxiv.org/abs/2109.05095}
\showURL{%
\tempurl}


\bibitem[\protect\citeauthoryear{Bhatt, Andrus, Weller, and Xiang}{Bhatt
  et~al\mbox{.}}{2020}]%
        {bhatt2020machine}
\bibfield{author}{\bibinfo{person}{Umang Bhatt}, \bibinfo{person}{McKane
  Andrus}, \bibinfo{person}{Adrian Weller}, {and} \bibinfo{person}{Alice
  Xiang}.} \bibinfo{year}{2020}\natexlab{}.
\newblock \showarticletitle{Machine learning explainability for external
  stakeholders}.
\newblock \bibinfo{journal}{\emph{arXiv preprint arXiv:2007.05408}}
  (\bibinfo{year}{2020}).
\newblock


\bibitem[\protect\citeauthoryear{Bhattacharyya, Phillips, Liu, Gupta,
  Driggs-Campbell, and Kochenderfer}{Bhattacharyya et~al\mbox{.}}{2019}]%
        {bhattacharyya2019simulating}
\bibfield{author}{\bibinfo{person}{Raunak~P Bhattacharyya},
  \bibinfo{person}{Derek~J Phillips}, \bibinfo{person}{Changliu Liu},
  \bibinfo{person}{Jayesh~K Gupta}, \bibinfo{person}{Katherine
  Driggs-Campbell}, {and} \bibinfo{person}{Mykel~J Kochenderfer}.}
  \bibinfo{year}{2019}\natexlab{}.
\newblock \showarticletitle{Simulating emergent properties of human driving
  behavior using multi-agent reward augmented imitation learning}. In
  \bibinfo{booktitle}{\emph{2019 International Conference on Robotics and
  Automation (ICRA)}}. IEEE, \bibinfo{pages}{789--795}.
\newblock


\bibitem[\protect\citeauthoryear{Bogdoll, Nitsche, and Z{\"o}llner}{Bogdoll
  et~al\mbox{.}}{2022}]%
        {bogdoll2022anomaly}
\bibfield{author}{\bibinfo{person}{Daniel Bogdoll}, \bibinfo{person}{Maximilian
  Nitsche}, {and} \bibinfo{person}{J~Marius Z{\"o}llner}.}
  \bibinfo{year}{2022}\natexlab{}.
\newblock \showarticletitle{Anomaly Detection in Autonomous Driving: A Survey}.
  In \bibinfo{booktitle}{\emph{Proceedings of the IEEE/CVF Conference on
  Computer Vision and Pattern Recognition}}. \bibinfo{pages}{4488--4499}.
\newblock


\bibitem[\protect\citeauthoryear{Boulanger-Lewandowski, Bengio, and
  Vincent}{Boulanger-Lewandowski et~al\mbox{.}}{2012}]%
        {bl2012modeling}
\bibfield{author}{\bibinfo{person}{Nicolas Boulanger-Lewandowski},
  \bibinfo{person}{Yoshua Bengio}, {and} \bibinfo{person}{Pascal Vincent}.}
  \bibinfo{year}{2012}\natexlab{}.
\newblock \showarticletitle{Modeling Temporal Dependencies in High-Dimensional
  Sequences: Application to Polyphonic Music Generation and Transcription}. In
  \bibinfo{booktitle}{\emph{Proceedings of the 29th International Coference on
  International Conference on Machine Learning}} (Edinburgh, Scotland)
  \emph{(\bibinfo{series}{ICML'12})}. \bibinfo{publisher}{Omnipress},
  \bibinfo{address}{Madison, WI, USA}, \bibinfo{pages}{1881–1888}.
\newblock
\showISBNx{9781450312851}


\bibitem[\protect\citeauthoryear{Chai, Sapp, Bansal, and Anguelov}{Chai
  et~al\mbox{.}}{2020}]%
        {chai2020multipath}
\bibfield{author}{\bibinfo{person}{Yuning Chai}, \bibinfo{person}{Benjamin
  Sapp}, \bibinfo{person}{Mayank Bansal}, {and} \bibinfo{person}{Dragomir
  Anguelov}.} \bibinfo{year}{2020}\natexlab{}.
\newblock \showarticletitle{MultiPath: Multiple Probabilistic Anchor Trajectory
  Hypotheses for Behavior Prediction}. In \bibinfo{booktitle}{\emph{Proceedings
  of the Conference on Robot Learning}} \emph{(\bibinfo{series}{Proceedings of
  Machine Learning Research}, Vol.~\bibinfo{volume}{100})},
  \bibfield{editor}{\bibinfo{person}{Leslie~Pack Kaelbling},
  \bibinfo{person}{Danica Kragic}, {and} \bibinfo{person}{Komei Sugiura}}
  (Eds.). \bibinfo{publisher}{PMLR}, \bibinfo{pages}{86--99}.
\newblock
\urldef\tempurl%
\url{https://proceedings.mlr.press/v100/chai20a.html}
\showURL{%
\tempurl}


\bibitem[\protect\citeauthoryear{Chandola, Banerjee, and Kumar}{Chandola
  et~al\mbox{.}}{2009}]%
        {chandola2009anomaly}
\bibfield{author}{\bibinfo{person}{Varun Chandola}, \bibinfo{person}{Arindam
  Banerjee}, {and} \bibinfo{person}{Vipin Kumar}.}
  \bibinfo{year}{2009}\natexlab{}.
\newblock \showarticletitle{Anomaly detection: A survey}.
\newblock \bibinfo{journal}{\emph{ACM computing surveys (CSUR)}}
  \bibinfo{volume}{41}, \bibinfo{number}{3} (\bibinfo{year}{2009}),
  \bibinfo{pages}{1--58}.
\newblock


\bibitem[\protect\citeauthoryear{Chen, Ding, Yang, Han, Xu, Gao, Zhang, Ouyang,
  Cai, and Chen}{Chen et~al\mbox{.}}{2021}]%
        {chen2021dual}
\bibfield{author}{\bibinfo{person}{Jingyuan Chen}, \bibinfo{person}{Guanchen
  Ding}, \bibinfo{person}{Yuchen Yang}, \bibinfo{person}{Wenwei Han},
  \bibinfo{person}{Kangmin Xu}, \bibinfo{person}{Tianyi Gao},
  \bibinfo{person}{Zhe Zhang}, \bibinfo{person}{Wanping Ouyang},
  \bibinfo{person}{Hao Cai}, {and} \bibinfo{person}{Zhenzhong Chen}.}
  \bibinfo{year}{2021}\natexlab{}.
\newblock \showarticletitle{Dual-modality vehicle anomaly detection via
  bilateral trajectory tracing}. In \bibinfo{booktitle}{\emph{Proceedings of
  the IEEE/CVF Conference on Computer Vision and Pattern Recognition}}.
  \bibinfo{pages}{4016--4025}.
\newblock


\bibitem[\protect\citeauthoryear{Choi, Lee, Jung, and Choi}{Choi
  et~al\mbox{.}}{2022}]%
        {choi2022multi}
\bibfield{author}{\bibinfo{person}{Taesung Choi}, \bibinfo{person}{Dongkun
  Lee}, \bibinfo{person}{Yuchae Jung}, {and} \bibinfo{person}{Ho-Jin Choi}.}
  \bibinfo{year}{2022}\natexlab{}.
\newblock \showarticletitle{Multivariate Time-series Anomaly Detection using
  SeqVAE-CNN Hybrid Model}. In \bibinfo{booktitle}{\emph{2022 International
  Conference on Information Networking (ICOIN)}}. \bibinfo{pages}{250--253}.
\newblock
\urldef\tempurl%
\url{https://doi.org/10.1109/ICOIN53446.2022.9687205}
\showDOI{\tempurl}


\bibitem[\protect\citeauthoryear{Chung, Kastner, Dinh, Goel, Courville, and
  Bengio}{Chung et~al\mbox{.}}{2015}]%
        {chung2015vrnn}
\bibfield{author}{\bibinfo{person}{Junyoung Chung}, \bibinfo{person}{Kyle
  Kastner}, \bibinfo{person}{Laurent Dinh}, \bibinfo{person}{Kratarth Goel},
  \bibinfo{person}{Aaron~C Courville}, {and} \bibinfo{person}{Yoshua Bengio}.}
  \bibinfo{year}{2015}\natexlab{}.
\newblock \showarticletitle{A Recurrent Latent Variable Model for Sequential
  Data}. In \bibinfo{booktitle}{\emph{Advances in Neural Information Processing
  Systems}}, \bibfield{editor}{\bibinfo{person}{C.~Cortes},
  \bibinfo{person}{N.~Lawrence}, \bibinfo{person}{D.~Lee},
  \bibinfo{person}{M.~Sugiyama}, {and} \bibinfo{person}{R.~Garnett}} (Eds.),
  Vol.~\bibinfo{volume}{28}. \bibinfo{publisher}{Curran Associates, Inc.}
\newblock
\urldef\tempurl%
\url{https://proceedings.neurips.cc/paper/2015/file/b618c3210e934362ac261db280128c22-Paper.pdf}
\showURL{%
\tempurl}


\bibitem[\protect\citeauthoryear{Costa, Sánchez, and Couso}{Costa
  et~al\mbox{.}}{2021}]%
        {costa2021semi}
\bibfield{author}{\bibinfo{person}{Nahuel Costa}, \bibinfo{person}{Luciano
  Sánchez}, {and} \bibinfo{person}{Inés Couso}.}
  \bibinfo{year}{2021}\natexlab{}.
\newblock \showarticletitle{Semi-Supervised Recurrent Variational Autoencoder
  Approach for Visual Diagnosis of Atrial Fibrillation}.
\newblock \bibinfo{journal}{\emph{IEEE Access}}  \bibinfo{volume}{9}
  (\bibinfo{year}{2021}), \bibinfo{pages}{40227--40239}.
\newblock
\urldef\tempurl%
\url{https://doi.org/10.1109/ACCESS.2021.3064854}
\showDOI{\tempurl}


\bibitem[\protect\citeauthoryear{De~Candido, Binder, and Utschick}{De~Candido
  et~al\mbox{.}}{2021}]%
        {Candido2021}
\bibfield{author}{\bibinfo{person}{Oliver De~Candido},
  \bibinfo{person}{Maximilian Binder}, {and} \bibinfo{person}{Wolfgang
  Utschick}.} \bibinfo{year}{2021}\natexlab{}.
\newblock \showarticletitle{An Interpretable Lane Change Detector Algorithm
  based on Deep Autoencoder Anomaly Detection}. In
  \bibinfo{booktitle}{\emph{2021 IEEE Intelligent Vehicles Symposium (IV)}}.
  \bibinfo{pages}{516--523}.
\newblock
\urldef\tempurl%
\url{https://doi.org/10.1109/IV48863.2021.9575599}
\showDOI{\tempurl}


\bibitem[\protect\citeauthoryear{Deo, Wolff, and Beijbom}{Deo
  et~al\mbox{.}}{2021}]%
        {deo2021multimodal}
\bibfield{author}{\bibinfo{person}{Nachiket Deo}, \bibinfo{person}{Eric Wolff},
  {and} \bibinfo{person}{Oscar Beijbom}.} \bibinfo{year}{2021}\natexlab{}.
\newblock \showarticletitle{Multimodal Trajectory Prediction Conditioned on
  Lane-Graph Traversals}. In \bibinfo{booktitle}{\emph{5th Annual Conference on
  Robot Learning}}.
\newblock
\urldef\tempurl%
\url{https://openreview.net/forum?id=hu7b7MPCqiC}
\showURL{%
\tempurl}


\bibitem[\protect\citeauthoryear{Dong, Chen, Peng, and Ma}{Dong
  et~al\mbox{.}}{2022}]%
        {supvssemisup}
\bibfield{author}{\bibinfo{person}{Yongqi Dong}, \bibinfo{person}{Kejia Chen},
  \bibinfo{person}{Yinxuan Peng}, {and} \bibinfo{person}{Zhiyuan Ma}.}
  \bibinfo{year}{2022}\natexlab{}.
\newblock \bibinfo{title}{Comparative Study on Supervised versus
  Semi-supervised Machine Learning for Anomaly Detection of In-vehicle CAN
  Network}.
\newblock
\newblock
\urldef\tempurl%
\url{https://doi.org/10.48550/ARXIV.2207.10286}
\showDOI{\tempurl}


\bibitem[\protect\citeauthoryear{Gregor, Danihelka, Graves, Rezende, and
  Wierstra}{Gregor et~al\mbox{.}}{2015}]%
        {gregor2015draw}
\bibfield{author}{\bibinfo{person}{Karol Gregor}, \bibinfo{person}{Ivo
  Danihelka}, \bibinfo{person}{Alex Graves}, \bibinfo{person}{Danilo Rezende},
  {and} \bibinfo{person}{Daan Wierstra}.} \bibinfo{year}{2015}\natexlab{}.
\newblock \showarticletitle{DRAW: A Recurrent Neural Network For Image
  Generation}. In \bibinfo{booktitle}{\emph{Proceedings of the 32nd
  International Conference on Machine Learning}}
  \emph{(\bibinfo{series}{Proceedings of Machine Learning Research},
  Vol.~\bibinfo{volume}{37})}, \bibfield{editor}{\bibinfo{person}{Francis Bach}
  {and} \bibinfo{person}{David Blei}} (Eds.). \bibinfo{publisher}{PMLR},
  \bibinfo{address}{Lille, France}, \bibinfo{pages}{1462--1471}.
\newblock
\urldef\tempurl%
\url{https://proceedings.mlr.press/v37/gregor15.html}
\showURL{%
\tempurl}


\bibitem[\protect\citeauthoryear{Han, Ellefsen, Li, Holmeset, and Zhang}{Han
  et~al\mbox{.}}{2021}]%
        {han2021fault}
\bibfield{author}{\bibinfo{person}{Peihua Han}, \bibinfo{person}{André~Listou
  Ellefsen}, \bibinfo{person}{Guoyuan Li}, \bibinfo{person}{Finn~Tore
  Holmeset}, {and} \bibinfo{person}{Houxiang Zhang}.}
  \bibinfo{year}{2021}\natexlab{}.
\newblock \showarticletitle{Fault Detection With LSTM-Based Variational
  Autoencoder for Maritime Components}.
\newblock \bibinfo{journal}{\emph{IEEE Sensors Journal}} \bibinfo{volume}{21},
  \bibinfo{number}{19} (\bibinfo{year}{2021}), \bibinfo{pages}{21903--21912}.
\newblock
\urldef\tempurl%
\url{https://doi.org/10.1109/JSEN.2021.3105226}
\showDOI{\tempurl}


\bibitem[\protect\citeauthoryear{Higgins, Matthey, Pal, Burgess, Glorot,
  Botvinick, Mohamed, and Lerchner}{Higgins et~al\mbox{.}}{2017}]%
        {Higgins2017betaVAELB}
\bibfield{author}{\bibinfo{person}{Irina Higgins}, \bibinfo{person}{Lo{\"i}c
  Matthey}, \bibinfo{person}{Arka Pal}, \bibinfo{person}{Christopher~P.
  Burgess}, \bibinfo{person}{Xavier Glorot}, \bibinfo{person}{Matthew~M.
  Botvinick}, \bibinfo{person}{Shakir Mohamed}, {and}
  \bibinfo{person}{Alexander Lerchner}.} \bibinfo{year}{2017}\natexlab{}.
\newblock \showarticletitle{beta-VAE: Learning Basic Visual Concepts with a
  Constrained Variational Framework}. In \bibinfo{booktitle}{\emph{ICLR}}.
\newblock


\bibitem[\protect\citeauthoryear{Ivanovic, Leung, Schmerling, and
  Pavone}{Ivanovic et~al\mbox{.}}{2020}]%
        {ivanovic2020multimodal}
\bibfield{author}{\bibinfo{person}{Boris Ivanovic}, \bibinfo{person}{Karen
  Leung}, \bibinfo{person}{Edward Schmerling}, {and} \bibinfo{person}{Marco
  Pavone}.} \bibinfo{year}{2020}\natexlab{}.
\newblock \showarticletitle{Multimodal deep generative models for trajectory
  prediction: A conditional variational autoencoder approach}.
\newblock \bibinfo{journal}{\emph{IEEE Robotics and Automation Letters}}
  \bibinfo{volume}{6}, \bibinfo{number}{2} (\bibinfo{year}{2020}),
  \bibinfo{pages}{295--302}.
\newblock


\bibitem[\protect\citeauthoryear{Ji, Sivakumar, Chowdhary, and
  Driggs-Campbell}{Ji et~al\mbox{.}}{2022}]%
        {ji2022proactive}
\bibfield{author}{\bibinfo{person}{Tianchen Ji},
  \bibinfo{person}{Arun~Narenthiran Sivakumar}, \bibinfo{person}{Girish
  Chowdhary}, {and} \bibinfo{person}{Katherine Driggs-Campbell}.}
  \bibinfo{year}{2022}\natexlab{}.
\newblock \showarticletitle{Proactive Anomaly Detection for Robot Navigation
  With Multi-Sensor Fusion}.
\newblock \bibinfo{journal}{\emph{IEEE Robotics and Automation Letters}}
  \bibinfo{volume}{7}, \bibinfo{number}{2} (\bibinfo{year}{2022}),
  \bibinfo{pages}{4975--4982}.
\newblock


\bibitem[\protect\citeauthoryear{Ji, Vuppala, Chowdhary, and
  Driggs-Campbell}{Ji et~al\mbox{.}}{2021}]%
        {ji2020multi}
\bibfield{author}{\bibinfo{person}{Tianchen Ji}, \bibinfo{person}{Sri~Theja
  Vuppala}, \bibinfo{person}{Girish Chowdhary}, {and}
  \bibinfo{person}{Katherine Driggs-Campbell}.}
  \bibinfo{year}{2021}\natexlab{}.
\newblock \showarticletitle{Multi-Modal Anomaly Detection for Unstructured and
  Uncertain Environments}. In \bibinfo{booktitle}{\emph{Conference on Robot
  Learning}}. \bibinfo{publisher}{PMLR}, \bibinfo{pages}{1443--1455}.
\newblock


\bibitem[\protect\citeauthoryear{Kindratenko, Mu, Zhan, Maloney, Hashemi, Rabe,
  Xu, Campbell, Peng, and Gropp}{Kindratenko et~al\mbox{.}}{2020}]%
        {HAL}
\bibfield{author}{\bibinfo{person}{Volodymyr Kindratenko},
  \bibinfo{person}{Dawei Mu}, \bibinfo{person}{Yan Zhan}, \bibinfo{person}{John
  Maloney}, \bibinfo{person}{Sayed~Hadi Hashemi}, \bibinfo{person}{Benjamin
  Rabe}, \bibinfo{person}{Ke Xu}, \bibinfo{person}{Roy Campbell},
  \bibinfo{person}{Jian Peng}, {and} \bibinfo{person}{William Gropp}.}
  \bibinfo{year}{2020}\natexlab{}.
\newblock \showarticletitle{HAL: Computer System for Scalable Deep Learning}.
  In \bibinfo{booktitle}{\emph{Practice and Experience in Advanced Research
  Computing}} (Portland, OR, USA) \emph{(\bibinfo{series}{PEARC '20})}.
  \bibinfo{publisher}{Association for Computing Machinery},
  \bibinfo{address}{New York, NY, USA}, \bibinfo{pages}{41–48}.
\newblock
\showISBNx{9781450366892}
\urldef\tempurl%
\url{https://doi.org/10.1145/3311790.3396649}
\showDOI{\tempurl}


\bibitem[\protect\citeauthoryear{Kingma and Welling}{Kingma and
  Welling}{2014}]%
        {kingma2014vae}
\bibfield{author}{\bibinfo{person}{D.P. Kingma} {and} \bibinfo{person}{M.
  Welling}.} \bibinfo{year}{2014}\natexlab{}.
\newblock \showarticletitle{Auto-encoding variational bayes.}. In
  \bibinfo{booktitle}{\emph{2nd International Conference on Learning
  Representations, ICLR 2014 - Conference Track Proceedings}}.
  \bibinfo{address}{Machine Learning Group, Universiteit van Amsterdam}.
\newblock


\bibitem[\protect\citeauthoryear{Lee, Choi, Vernaza, Choy, Torr, and
  Chandraker}{Lee et~al\mbox{.}}{2017}]%
        {lee2017desire}
\bibfield{author}{\bibinfo{person}{Namhoon Lee}, \bibinfo{person}{Wongun Choi},
  \bibinfo{person}{Paul Vernaza}, \bibinfo{person}{Christopher~B. Choy},
  \bibinfo{person}{Philip H.~S. Torr}, {and} \bibinfo{person}{Manmohan
  Chandraker}.} \bibinfo{year}{2017}\natexlab{}.
\newblock \showarticletitle{DESIRE: Distant Future Prediction in Dynamic Scenes
  with Interacting Agents}. In \bibinfo{booktitle}{\emph{2017 IEEE Conference
  on Computer Vision and Pattern Recognition (CVPR)}}.
  \bibinfo{pages}{2165--2174}.
\newblock
\urldef\tempurl%
\url{https://doi.org/10.1109/CVPR.2017.233}
\showDOI{\tempurl}


\bibitem[\protect\citeauthoryear{Liang, Yang, Hu, Chen, Liao, Feng, and
  Urtasun}{Liang et~al\mbox{.}}{2020}]%
        {LANEGCN2020}
\bibfield{author}{\bibinfo{person}{Ming Liang}, \bibinfo{person}{Bin Yang},
  \bibinfo{person}{Rui Hu}, \bibinfo{person}{Yun Chen}, \bibinfo{person}{Renjie
  Liao}, \bibinfo{person}{Song Feng}, {and} \bibinfo{person}{Raquel Urtasun}.}
  \bibinfo{year}{2020}\natexlab{}.
\newblock \showarticletitle{Learning Lane Graph Representations for Motion
  Forecasting}. In \bibinfo{booktitle}{\emph{Computer Vision -- ECCV 2020}},
  \bibfield{editor}{\bibinfo{person}{Andrea Vedaldi}, \bibinfo{person}{Horst
  Bischof}, \bibinfo{person}{Thomas Brox}, {and} \bibinfo{person}{Jan-Michael
  Frahm}} (Eds.). \bibinfo{publisher}{Springer International Publishing},
  \bibinfo{address}{Cham}, \bibinfo{pages}{541--556}.
\newblock
\showISBNx{978-3-030-58536-5}


\bibitem[\protect\citeauthoryear{Liu, Chang, Chen, Chakraborty, and
  Driggs-Campbell}{Liu et~al\mbox{.}}{2022}]%
        {liu2022learning}
\bibfield{author}{\bibinfo{person}{Shuijing Liu}, \bibinfo{person}{Peixin
  Chang}, \bibinfo{person}{Haonan Chen}, \bibinfo{person}{Neeloy Chakraborty},
  {and} \bibinfo{person}{Katherine Driggs-Campbell}.}
  \bibinfo{year}{2022}\natexlab{}.
\newblock \showarticletitle{Learning to Navigate Intersections with
  Unsupervised Driver Trait Inference}. In \bibinfo{booktitle}{\emph{2022
  International Conference on Robotics and Automation (ICRA)}}.
  \bibinfo{pages}{3576--3582}.
\newblock
\urldef\tempurl%
\url{https://doi.org/10.1109/ICRA46639.2022.9811635}
\showDOI{\tempurl}


\bibitem[\protect\citeauthoryear{Mi, Zhao, Nash, Jin, Gao, Sun, Schmid, Shavit,
  Chai, and Anguelov}{Mi et~al\mbox{.}}{2021}]%
        {mi2021hdmapgen}
\bibfield{author}{\bibinfo{person}{Lu Mi}, \bibinfo{person}{Hang Zhao},
  \bibinfo{person}{Charlie Nash}, \bibinfo{person}{Xiaohan Jin},
  \bibinfo{person}{Jiyang Gao}, \bibinfo{person}{Chen Sun},
  \bibinfo{person}{Cordelia Schmid}, \bibinfo{person}{Nir Shavit},
  \bibinfo{person}{Yuning Chai}, {and} \bibinfo{person}{Dragomir Anguelov}.}
  \bibinfo{year}{2021}\natexlab{}.
\newblock \showarticletitle{HDMapGen: A Hierarchical Graph Generative Model of
  High Definition Maps}. In \bibinfo{booktitle}{\emph{Proceedings of the
  IEEE/CVF Conference on Computer Vision and Pattern Recognition (CVPR)}}.
  \bibinfo{pages}{4227--4236}.
\newblock


\bibitem[\protect\citeauthoryear{Morton, Witherden, and Kochenderfer}{Morton
  et~al\mbox{.}}{2019}]%
        {ijcai2019p440}
\bibfield{author}{\bibinfo{person}{Jeremy Morton}, \bibinfo{person}{Freddie~D.
  Witherden}, {and} \bibinfo{person}{Mykel~J. Kochenderfer}.}
  \bibinfo{year}{2019}\natexlab{}.
\newblock \showarticletitle{Deep Variational Koopman Models: Inferring Koopman
  Observations for Uncertainty-Aware Dynamics Modeling and Control}. In
  \bibinfo{booktitle}{\emph{Proceedings of the Twenty-Eighth International
  Joint Conference on Artificial Intelligence, {IJCAI-19}}}.
  \bibinfo{publisher}{International Joint Conferences on Artificial
  Intelligence Organization}, \bibinfo{pages}{3173--3179}.
\newblock
\urldef\tempurl%
\url{https://doi.org/10.24963/ijcai.2019/440}
\showDOI{\tempurl}


\bibitem[\protect\citeauthoryear{Pang, Shen, Cao, and Hengel}{Pang
  et~al\mbox{.}}{2021}]%
        {pang2021deep}
\bibfield{author}{\bibinfo{person}{Guansong Pang}, \bibinfo{person}{Chunhua
  Shen}, \bibinfo{person}{Longbing Cao}, {and} \bibinfo{person}{Anton Van~Den
  Hengel}.} \bibinfo{year}{2021}\natexlab{}.
\newblock \showarticletitle{Deep learning for anomaly detection: A review}.
\newblock \bibinfo{journal}{\emph{ACM Computing Surveys (CSUR)}}
  \bibinfo{volume}{54}, \bibinfo{number}{2} (\bibinfo{year}{2021}),
  \bibinfo{pages}{1--38}.
\newblock


\bibitem[\protect\citeauthoryear{Park, Erickson, Bhattacharjee, and Kemp}{Park
  et~al\mbox{.}}{2016}]%
        {park2016multimodal}
\bibfield{author}{\bibinfo{person}{Daehyung Park}, \bibinfo{person}{Zackory
  Erickson}, \bibinfo{person}{Tapomayukh Bhattacharjee}, {and}
  \bibinfo{person}{Charles~C Kemp}.} \bibinfo{year}{2016}\natexlab{}.
\newblock \showarticletitle{Multimodal execution monitoring for anomaly
  detection during robot manipulation}. In \bibinfo{booktitle}{\emph{2016 IEEE
  International Conference on Robotics and Automation (ICRA)}}. IEEE,
  \bibinfo{pages}{407--414}.
\newblock


\bibitem[\protect\citeauthoryear{Park, Hoshi, and Kemp}{Park
  et~al\mbox{.}}{2018}]%
        {park2018multimodal}
\bibfield{author}{\bibinfo{person}{Daehyung Park}, \bibinfo{person}{Yuuna
  Hoshi}, {and} \bibinfo{person}{Charles~C Kemp}.}
  \bibinfo{year}{2018}\natexlab{}.
\newblock \showarticletitle{A multimodal anomaly detector for robot-assisted
  feeding using an lstm-based variational autoencoder}.
\newblock \bibinfo{journal}{\emph{IEEE Robotics and Automation Letters}}
  \bibinfo{volume}{3}, \bibinfo{number}{3} (\bibinfo{year}{2018}),
  \bibinfo{pages}{1544--1551}.
\newblock


\bibitem[\protect\citeauthoryear{Salzmann, Ivanovic, Chakravarty, and
  Pavone}{Salzmann et~al\mbox{.}}{2020}]%
        {salzmann2020trajectron++}
\bibfield{author}{\bibinfo{person}{Tim Salzmann}, \bibinfo{person}{Boris
  Ivanovic}, \bibinfo{person}{Punarjay Chakravarty}, {and}
  \bibinfo{person}{Marco Pavone}.} \bibinfo{year}{2020}\natexlab{}.
\newblock \showarticletitle{Trajectron++: Dynamically-feasible trajectory
  forecasting with heterogeneous data}. In \bibinfo{booktitle}{\emph{European
  Conference on Computer Vision}}. Springer, \bibinfo{pages}{683--700}.
\newblock


\bibitem[\protect\citeauthoryear{Sch{\"o}ller, Aravantinos, Lay, and
  Knoll}{Sch{\"o}ller et~al\mbox{.}}{2020}]%
        {scholler2020constant}
\bibfield{author}{\bibinfo{person}{Christoph Sch{\"o}ller},
  \bibinfo{person}{Vincent Aravantinos}, \bibinfo{person}{Florian Lay}, {and}
  \bibinfo{person}{Alois Knoll}.} \bibinfo{year}{2020}\natexlab{}.
\newblock \showarticletitle{What the constant velocity model can teach us about
  pedestrian motion prediction}.
\newblock \bibinfo{journal}{\emph{IEEE Robotics and Automation Letters}}
  \bibinfo{volume}{5}, \bibinfo{number}{2} (\bibinfo{year}{2020}),
  \bibinfo{pages}{1696--1703}.
\newblock


\bibitem[\protect\citeauthoryear{Sejr and Schneider-Kamp}{Sejr and
  Schneider-Kamp}{2021}]%
        {SEJR2021}
\bibfield{author}{\bibinfo{person}{Jonas~Herskind Sejr} {and}
  \bibinfo{person}{Anna Schneider-Kamp}.} \bibinfo{year}{2021}\natexlab{}.
\newblock \showarticletitle{Explainable outlier detection: What, for Whom and
  Why?}
\newblock \bibinfo{journal}{\emph{Machine Learning with Applications}}
  \bibinfo{volume}{6} (\bibinfo{year}{2021}), \bibinfo{pages}{100172}.
\newblock
\showISSN{2666-8270}
\urldef\tempurl%
\url{https://doi.org/10.1016/j.mlwa.2021.100172}
\showDOI{\tempurl}


\bibitem[\protect\citeauthoryear{Shah, ling Huang, Laddha, Langford, Barber,
  Zhang, Vallespi-Gonzalez, and Urtasun}{Shah et~al\mbox{.}}{2020}]%
        {Shah2020LiRaNetET}
\bibfield{author}{\bibinfo{person}{Meet Shah}, \bibinfo{person}{Zhi ling
  Huang}, \bibinfo{person}{Ankita~Gajanan Laddha}, \bibinfo{person}{Matthew
  Langford}, \bibinfo{person}{Blake Barber}, \bibinfo{person}{Sidney Zhang},
  \bibinfo{person}{Carlos Vallespi-Gonzalez}, {and} \bibinfo{person}{Raquel
  Urtasun}.} \bibinfo{year}{2020}\natexlab{}.
\newblock \showarticletitle{LiRaNet: End-to-End Trajectory Prediction using
  Spatio-Temporal Radar Fusion}. In \bibinfo{booktitle}{\emph{CoRL}}.
\newblock


\bibitem[\protect\citeauthoryear{Sipple and Youssef}{Sipple and
  Youssef}{2022}]%
        {sipple2022general}
\bibfield{author}{\bibinfo{person}{John Sipple} {and} \bibinfo{person}{Abdou
  Youssef}.} \bibinfo{year}{2022}\natexlab{}.
\newblock \showarticletitle{A general-purpose method for applying Explainable
  AI for Anomaly Detection}. In \bibinfo{booktitle}{\emph{International
  Symposium on Methodologies for Intelligent Systems}}. Springer,
  \bibinfo{pages}{162--174}.
\newblock


\bibitem[\protect\citeauthoryear{Sohn, Lee, and Yan}{Sohn
  et~al\mbox{.}}{2015}]%
        {NIPS2015_8d55a249}
\bibfield{author}{\bibinfo{person}{Kihyuk Sohn}, \bibinfo{person}{Honglak Lee},
  {and} \bibinfo{person}{Xinchen Yan}.} \bibinfo{year}{2015}\natexlab{}.
\newblock \showarticletitle{Learning Structured Output Representation using
  Deep Conditional Generative Models}. In \bibinfo{booktitle}{\emph{Advances in
  Neural Information Processing Systems}},
  \bibfield{editor}{\bibinfo{person}{C.~Cortes}, \bibinfo{person}{N.~Lawrence},
  \bibinfo{person}{D.~Lee}, \bibinfo{person}{M.~Sugiyama}, {and}
  \bibinfo{person}{R.~Garnett}} (Eds.), Vol.~\bibinfo{volume}{28}.
  \bibinfo{publisher}{Curran Associates, Inc.}
\newblock
\urldef\tempurl%
\url{https://proceedings.neurips.cc/paper/2015/file/8d55a249e6baa5c06772297520da2051-Paper.pdf}
\showURL{%
\tempurl}


\bibitem[\protect\citeauthoryear{Vaswani, Shazeer, Parmar, Uszkoreit, Jones,
  Gomez, Kaiser, and Polosukhin}{Vaswani et~al\mbox{.}}{2017}]%
        {attention}
\bibfield{author}{\bibinfo{person}{Ashish Vaswani}, \bibinfo{person}{Noam
  Shazeer}, \bibinfo{person}{Niki Parmar}, \bibinfo{person}{Jakob Uszkoreit},
  \bibinfo{person}{Llion Jones}, \bibinfo{person}{Aidan~N. Gomez},
  \bibinfo{person}{Lukasz Kaiser}, {and} \bibinfo{person}{Illia Polosukhin}.}
  \bibinfo{year}{2017}\natexlab{}.
\newblock \bibinfo{title}{Attention Is All You Need}.
\newblock
\newblock
\urldef\tempurl%
\url{https://doi.org/10.48550/ARXIV.1706.03762}
\showDOI{\tempurl}


\bibitem[\protect\citeauthoryear{Wiederer, Bouazizi, Troina, Kressel, and
  Belagiannis}{Wiederer et~al\mbox{.}}{2022}]%
        {MAAD2021}
\bibfield{author}{\bibinfo{person}{Julian Wiederer}, \bibinfo{person}{Arij
  Bouazizi}, \bibinfo{person}{Marco Troina}, \bibinfo{person}{Ulrich Kressel},
  {and} \bibinfo{person}{Vasileios Belagiannis}.}
  \bibinfo{year}{2022}\natexlab{}.
\newblock \showarticletitle{Anomaly Detection in Multi-Agent Trajectories for
  Automated Driving}. In \bibinfo{booktitle}{\emph{Conference on Robot
  Learning}}. PMLR, \bibinfo{pages}{1223--1233}.
\newblock


\bibitem[\protect\citeauthoryear{Yang, Renzaglia, Paigwar, Laugier, and
  Wang}{Yang et~al\mbox{.}}{2019}]%
        {YangDrivingBeh}
\bibfield{author}{\bibinfo{person}{Chule Yang}, \bibinfo{person}{Alessandro
  Renzaglia}, \bibinfo{person}{Anshul Paigwar}, \bibinfo{person}{Christian
  Laugier}, {and} \bibinfo{person}{Danwei Wang}.}
  \bibinfo{year}{2019}\natexlab{}.
\newblock \showarticletitle{Driving Behavior Assessment and Anomaly Detection
  for Intelligent Vehicles}. In \bibinfo{booktitle}{\emph{2019 IEEE
  International Conference on Cybernetics and Intelligent Systems (CIS) and
  IEEE Conference on Robotics, Automation and Mechatronics (RAM)}}.
  \bibinfo{pages}{524--529}.
\newblock
\urldef\tempurl%
\url{https://doi.org/10.1109/CIS-RAM47153.2019.9095790}
\showDOI{\tempurl}


\bibitem[\protect\citeauthoryear{Yao, Wang, Xu, Pu, Wang, Atkins, and
  Crandall}{Yao et~al\mbox{.}}{2022}]%
        {yao2022dota}
\bibfield{author}{\bibinfo{person}{Yu Yao}, \bibinfo{person}{Xizi Wang},
  \bibinfo{person}{Mingze Xu}, \bibinfo{person}{Zelin Pu},
  \bibinfo{person}{Yuchen Wang}, \bibinfo{person}{Ella Atkins}, {and}
  \bibinfo{person}{David Crandall}.} \bibinfo{year}{2022}\natexlab{}.
\newblock \showarticletitle{DoTA: unsupervised detection of traffic anomaly in
  driving videos}.
\newblock \bibinfo{journal}{\emph{IEEE transactions on pattern analysis and
  machine intelligence}} (\bibinfo{year}{2022}).
\newblock


\bibitem[\protect\citeauthoryear{Yao, Xu, Wang, Crandall, and Atkins}{Yao
  et~al\mbox{.}}{2019}]%
        {yao2019unsupervised}
\bibfield{author}{\bibinfo{person}{Yu Yao}, \bibinfo{person}{Mingze Xu},
  \bibinfo{person}{Yuchen Wang}, \bibinfo{person}{David~J Crandall}, {and}
  \bibinfo{person}{Ella~M Atkins}.} \bibinfo{year}{2019}\natexlab{}.
\newblock \showarticletitle{Unsupervised traffic accident detection in
  first-person videos}. In \bibinfo{booktitle}{\emph{2019 IEEE/RSJ
  International Conference on Intelligent Robots and Systems (IROS)}}. IEEE,
  \bibinfo{pages}{273--280}.
\newblock


\bibitem[\protect\citeauthoryear{Yoon, Jeong, and Sung}{Yoon
  et~al\mbox{.}}{2022}]%
        {Yoon2022design}
\bibfield{author}{\bibinfo{person}{Jun~Yong Yoon}, \bibinfo{person}{Jinseop
  Jeong}, {and} \bibinfo{person}{Woosuk Sung}.}
  \bibinfo{year}{2022}\natexlab{}.
\newblock \showarticletitle{Design and Implementation of HD Mapping, Vehicle
  Control, and V2I Communication for Robo-Taxi Services}.
\newblock \bibinfo{journal}{\emph{Sensors}} \bibinfo{volume}{22},
  \bibinfo{number}{18} (\bibinfo{year}{2022}).
\newblock
\urldef\tempurl%
\url{https://doi.org/10.3390/s22187049}
\showDOI{\tempurl}


\bibitem[\protect\citeauthoryear{Zhang, Tao, Tan, T{\"o}rngren, S{\'a}nchez,
  Ramli, Tao, Gyllenhammar, Wotawa, Mohan, Nica, and Felbinger}{Zhang
  et~al\mbox{.}}{2021}]%
        {Zhang2021FindingCS}
\bibfield{author}{\bibinfo{person}{Xinhai Zhang}, \bibinfo{person}{Jianbo Tao},
  \bibinfo{person}{Kaige Tan}, \bibinfo{person}{Martin T{\"o}rngren},
  \bibinfo{person}{Jos{\'e} Manuel~Gaspar S{\'a}nchez},
  \bibinfo{person}{Muhammad~Rusyadi Ramli}, \bibinfo{person}{Xin Tao},
  \bibinfo{person}{Magnus Gyllenhammar}, \bibinfo{person}{Franz Wotawa},
  \bibinfo{person}{Naveen Mohan}, \bibinfo{person}{Mihai Nica}, {and}
  \bibinfo{person}{Hermann Felbinger}.} \bibinfo{year}{2021}\natexlab{}.
\newblock \showarticletitle{Finding Critical Scenarios for Automated Driving
  Systems: A Systematic Literature Review}.
\newblock \bibinfo{journal}{\emph{ArXiv}}  \bibinfo{volume}{abs/2110.08664}
  (\bibinfo{year}{2021}).
\newblock


\end{thebibliography}


\newpage
\onecolumn
\section*{Appendix}

\appendix

\section{Training and Hyperparameters}
\label{sec:app_train}
Several runs of SABeR-VAE were trained with learning rates ranging from $5e-5$ to $1e-3$, batch sizes from $32$ to $128$, KL-Divergence $\beta_1=\beta_2$ weighting from $1e-6$ to $1e-3$, latent dimension size from $2$ to $64$, attention embedding sizes $32$ and $64$, a weight decay of $1e-6$, and an inter-vehicle distance $d$ of $45$ meters for the vehicle-vehicle attention mask. The RAE-Pred, VV-RAE, SABeR-AE, and Att-LSTM-VAE models were similarly trained, but without $\beta$, attention embedding sizes, and $d$ where irrelevant. Multi-head attention modules were instantiated with $8$ heads each. Every model was trained for $500$ epochs on the training split with a Tesla V100 GPU. 

\begin{table}[ht]
\centering
\caption{\textbf{Finalized hyperparameters per method.}}
\label{tab:hyperparams}
\begin{tabular}{lccccc}
\toprule
\multicolumn{1}{c}{Hyperparameter} & RAE-Pred & VV-RAE & Att-LSTM-VAE & SABeR-AE & SABeR-VAE \\
\midrule
Learning Rate                      & $5e-5$   & $5e-4$ & $5e-5$       & $5e-5$   & $5e-5$    \\
Batch Size                         & $32$     & $128$  & $32$         & $64$     & $32$      \\
GRU Encoder Embedding Size         & $64$     & $32$   & $32$         & $64$     & $32$      \\
Latent Dimension Size              & $2$      & $2$    & $2$          & $64$     & $2$       \\
Attention Embedding Sizes          &          & $32$   & $32$         & $64$     & $32$      \\
KL-Divergence Beta                 &          &        & $1e-6$       &          & $1e-6$    \\
\bottomrule
\end{tabular}
\end{table}

\section{AUROC by Anomaly Type}
\label{sec:app_auroc}

\begin{table*}[ht]
\centering
\caption{\textbf{AUROC ($\uparrow$) of methods by anomaly type.}}
\label{tab:anom_accuracy}
\begin{threeparttable}
\begin{tabular}{lcccccc}
\toprule
Anomaly Type                     & CVM  & STGAE-KDE\tnote{*} & RAE-Pred & VV-RAE & SABeR-AE & SABeR-VAE \\ 
\midrule
Pushing Aside                    & \textbf{91.5} & 90.4               & 87.1     & 54.5   & 88.6                 & 91.3     \\
Right Spreading                  & 87.7          & \textbf{96.2}      & 90.2     & 55.3   & 86.7                 & 95.2     \\
Left Spreading                   & 90.9          & \textbf{96.6}      & 93.4     & 60.0   & 96.2                 & 95.9     \\
Off-Road                         & 88.7          & 98.2               & 83.4     & 59.5   & 98.2                 & \textbf{99.7}     \\
Skidding                         & 96.9          & 99.7               & $\sim$\textbf{100.0}      & 69.6   & $\sim$\textbf{100.0} & 99.8     \\
Wrong-way                        & 63.2          & 73.2               & 67.8     & 73.3   & $\sim$\textbf{100.0} & 99.3     \\ 
Tailgating                       & 77.2          & \textbf{84.9}      & 75.9     & 42.1   & 74.8                 & 78.8     \\
Staggering                       & 86.9          & \textbf{96.0}      & 91.9     & 68.4   & 93.2                 & 92.5     \\
Reeving                          & \textbf{96.7} & 94.6               & 91.1     & 54.2   & 87.9                 & 89.5     \\
Aggressive Overtaking            & 91.8          & \textbf{93.2}               & 86.3     & 47.9   & 80.8                 & 77.9     \\
Thwarting                        & \textbf{98.1}          & 81.6               & 92.8     & 60.9   & 96.4                 & 93.9     \\
\midrule
Average                          & 88.1          & 91.3               & 87.3     & 58.7   & 91.2                 & \textbf{92.2}     \\
\bottomrule
\end{tabular}
\begin{tablenotes}
\item[*] These results are presented in~\cite{MAAD2021}.
\end{tablenotes}
\end{threeparttable}
\end{table*}

\end{document}